\documentclass[runningheads, envcountsame, a4paper]{llncs}

\usepackage[pdftex]{graphicx}
\usepackage{blindtext}
\usepackage{soul}
\usepackage{xcolor}
\usepackage[hyphens]{url}
\usepackage{amsmath,amsfonts,amsthm,bm} 
\usepackage{booktabs}
\usepackage{multirow}
\usepackage{enumerate}
\usepackage[linesnumbered,ruled]{algorithm2e}
\usepackage{pifont}
\usepackage[acronym]{glossaries}
\usepackage{wrapfig}
\usepackage[misc]{ifsym}
\usepackage{filecontents}

\newcommand{\rpm}{\raisebox{.2ex}{$\scriptstyle\pm$}}

\newcommand{\m}[1]{\ensuremath{\mathrm{#1}}}
\newcommand{\F}{\ensuremath{\mathcal{F}}}

\renewcommand{\arraystretch}{0.7}

\SetCommentSty{mycommfont}
\SetKwInput{KwInput}{Require}                
\SetKwInput{KwOutput}{Output}              

\graphicspath{ {./figures/} }

\newacronym{method}{SREA}{Self-Re-Labeling with Embedding Analysis}
\newacronym{nilm}{NILM}{Non-Intrusive Load Monitoring}

\begin{document}

\title{Estimating the Electrical Power Output of Industrial Devices with End-to-End Time-Series Classification in the Presence of Label Noise }

\titlerunning{Estimating Power Output of Industrial Devices in the Presence of Label Noise}

\toctitle{Estimating the Electrical Power Output of Industrial Devices with End-to-End Time-Series Classification in the Presence of Label Noise}
\tocauthor{Andrea~Castellani, Sebastian~Schmitt, Barbara~Hammer}

\newcommand{\orcidauthorA}{\orcidID{0000-0003-0476-5978}} 
\newcommand{\orcidauthorB}{\orcidID{0000-0001-7130-5483}} 
\newcommand{\orcidauthorC}{\orcidID{0000-0002-0935-5591}} 

\author{Andrea~Castellani (\Letter) \inst{1} \orcidauthorA  \and
Sebastian~Schmitt \inst{2} \orcidauthorB \and
Barbara~Hammer \inst{1} \orcidauthorC} 
\authorrunning{A. Castellani and S. Schmitt and B. Hammer}
%
\institute{Bielefeld University, \\
\email{\{acastellani,bhammer\}@techfak.uni-bielefeld.de}\\ 
\and
Honda Research Institute Europe, \\
\email{sebastian.schmitt@honda-ri.de}
}
\maketitle              
\begin{abstract}
In complex industrial settings, it is common practice to monitor the operation of machines in order to detect undesired states, adjust maintenance schedules, optimize system performance or collect usage statistics of individual machines. 
In this work, we focus on estimating the power output of a Combined Heat and Power (CHP) machine of a medium-sized company facility by analyzing the total facility power consumption.
We formulate the problem as a time-series classification problem, where the class label represents the CHP power output. 
As the facility is fully instrumented and sensor measurements from the CHP are available, we generate the training labels in an automated fashion from the CHP sensor readings. 
However, sensor failures result in mislabeled training data samples which are hard to detect and remove from the dataset. 
Therefore, we propose a novel multi-task deep learning approach that jointly trains a classifier and an autoencoder with a shared embedding representation.
The proposed approach targets to gradually correct the mislabelled data samples during training in a self-supervised fashion, without any prior assumption on the amount of label noise.
We benchmark our approach on several time-series classification datasets and find it to be comparable and sometimes better than state-of-the-art methods.
On the real-world use-case of predicting the CHP power output, we thoroughly evaluate the architectural design choices and show that the final architecture considerably increases the robustness of the learning process and consistently beats other recent state-of-the-art algorithms in the presence of unstructured as well as structured label noise.

\keywords{Time-series \and Deep Learning \and Label noise \and Self-supervision \and \acrlong{nilm} \and Time-Series Classification}
\end{abstract}
\section{Introduction}
It is common to monitor multiple machines in complex industrial settings for many diverse reasons, such as to detect undesired operational states, adjust maintenance schedules or optimize system performance.
In situations where the installation of many sensors for individual devices is not feasible due to cost or technical reasons, \acrfull{nilm} \cite{Holmegaard2016NILMIA} is able to identify the utilization of individual machines based on the analysis of cumulative electrical load profiles. 
The problem of the generation of labelled training data sets is a cornerstone of data-driven approaches to \acrshort{nilm}.
In this context,  industry  relies mostly on manually annotated data \cite{Fredriksson2020DataLA} and less often the training labels can be automatically generated from the sensors \cite{Gan2018AutomaticLF}, which is often unreliable because of sensor failure and  human misinterpretation.
Data cleaning techniques are often hard to implement \cite{Wang2020TimeSD} which unavoidably leads to the presence of wrongly annotated instances in automatically generated datasets i.e.\ \textit{label noise} \cite{frenay2013classification}. 
Many machine learning methods, and in particular deep neural networks, are able to overfit  training data with noisy labels \cite{zhang2016understanding}, thus it is challenging to apply data-driven approaches successfully in complex industrial settings.

We consider a medium-sized company facility and
 target the problem of estimating the electrical power output of a Combined Heat and Power (CHP) machine by only analyzing the facility electrical power consumption.
The electrical power output of the CHP is sufficient to supply a substantial share of the total electricity demand of the facility.
Therefore, knowing the CHP's electrical power output is very helpful for distributing the electrical energy in the facility, for example when scheduling the charging of electrical vehicles (EVs) or reducing total peak-load \cite{limmerEVCharging2019}.
We propose a data-driven deep learning-based approach to this problem, which is modelled as time-series classification challenge in the presence of label noise, where the class label of each time series represents the estimated CHP power output level.
As the facility is fully instrumented, and sensor measurements from the CHP are available, we generate the training labels in an automated fashion from the CHP sensor readings. 
However, these sensors fail from time to time which resulting wrong labels.
To tackle this problem, we propose a novel multi-task deep learning approach named \acrfull{method}, which targets the detection and re-labeling of wrongly labeled instances in a self-supervised training fashion.

In the following, after a formal introduction to \acrshort{method}, we empirically validate it with several benchmarks data sets for time-series classification and compare against state-of-the-art (SotA) algorithms.
In order to evaluate the performance, we create a training data set from clean sensor readings 
and we corrupt it by introducing three types of artificial noise in a controlled fashion.
After that, we apply the proposed method to a real-world use-case including real sensor failures and show that those are properly detected and corrected by our algorithm.  
Finally, we perform an extensive ablation study to investigate the sensitivity of the \acrshort{method} to its hyper-parameters.

\section{Related work}
Deep learning based techniques constitute a promising approach to solve the NILM problem \cite{massidda2020non,Paresh2020MultiLabelAB}. 
Since the requirement of large annotated data sets is very challenging, the problem is often addressed as a semi-supervised learning task
\cite{humala2018universalnilm,barsim2015toward,yang2019semisupervised}, where only a part of the data is correctly labeled, and the other is left without any label.
Here we take a different stance to the problem of energy disaggregation: we assume that labels for all data are given, but not all labels are  correct, as is often the case in complex sensor networks \cite{frenay2013classification}.

Learning noisy labels is an important topic in machine learning research  \cite{song2020learning,han2020survey}.
Several approaches are based on the fact that deep neural networks tend to first learn clean data statistics during early stages of training  \cite{rolnick2017deep,zhang2016understanding,arpit2017closer}.
Methods can be based on a loss function which is robust to label noise \cite{zhang2018generalized,van2015learning}, or they can introduce an explicit \cite{berthelot2019mixmatch} or implicit \cite{reed2015training} regularization term.
Another adaptation of the loss function is achieved by using different bootstrapping methods based on the predicted class labels \cite{han2020survey,nguyen2019self,wang2020self}.
Some other common approaches are based on labeling samples with smaller loss as clean ones \cite{arazo2019unsupervised,han2020sigua,jiang2018mentornet,li2020dividemix}, and fitting a two-component mixture model on the loss distribution, or using cross-validation techniques \cite{chen2019understanding} in order to separate clean and mislabeled samples.
Several existing methods \cite{sugiyama2018co,mandal2020novel,han2020sigua} require knowledge about the percentage of noisy samples, or estimate this quantity based on  a given. The suitability of these approaches is unclear for real-world applications.
Since our proposed method, \acrshort{method}, targets the correction of mislabeled training samples based on their embedding representation, we do not rely on specific  assumptions on the  amount of label noise in advance.
Our modeling approach is based on the observation that self-supervision 
has proven to provide an effective representation for downstream tasks without requiring labels \cite{hendrycks2019using,huang2021self}, this leading to an improvement of performance in the main supervised task \cite{jawed2020self}.

Most applications of models which deal with label noise come from the domain of image data, where noise can be induced by e.g.\ crowd-working annotations  \cite{Karimi2020DeepLW,han2020survey}. Some approaches consider label noise in other domains such as human activity detection \cite{atkinson2020identifying}, sound event classification \cite{fonseca2019learning}, or malware detection \cite{Gavrilut2011DealingWC}.
An attempt to analyze the effect of noisy data on applications for real-world data sets is made in \cite{Wang2019ALN}, but the authors do not compare to the SotA.
Up to the authors knowledge, we are the first to report a detailed evaluation of time-series classification in the presence of label noise, which is a crucial data characteristics in the domain of \acrshort{nilm}.

\section{CHP electrical power output estimation}\label{sec:CHP}
\begin{figure}[tbh]
    \centering
    \includegraphics[width=0.75\linewidth]{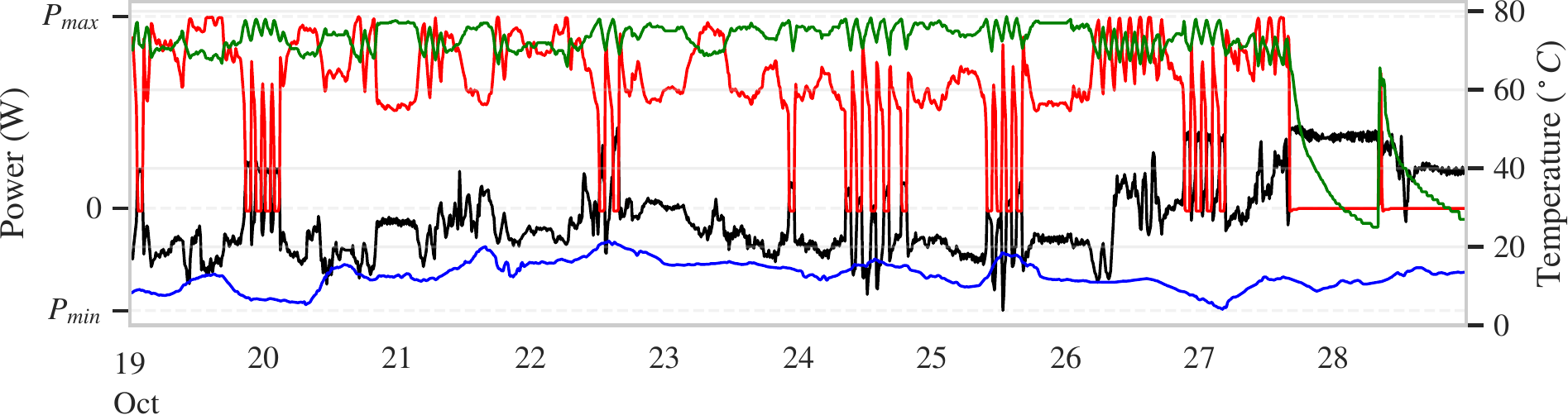}\\
    \includegraphics[width=0.75\linewidth]{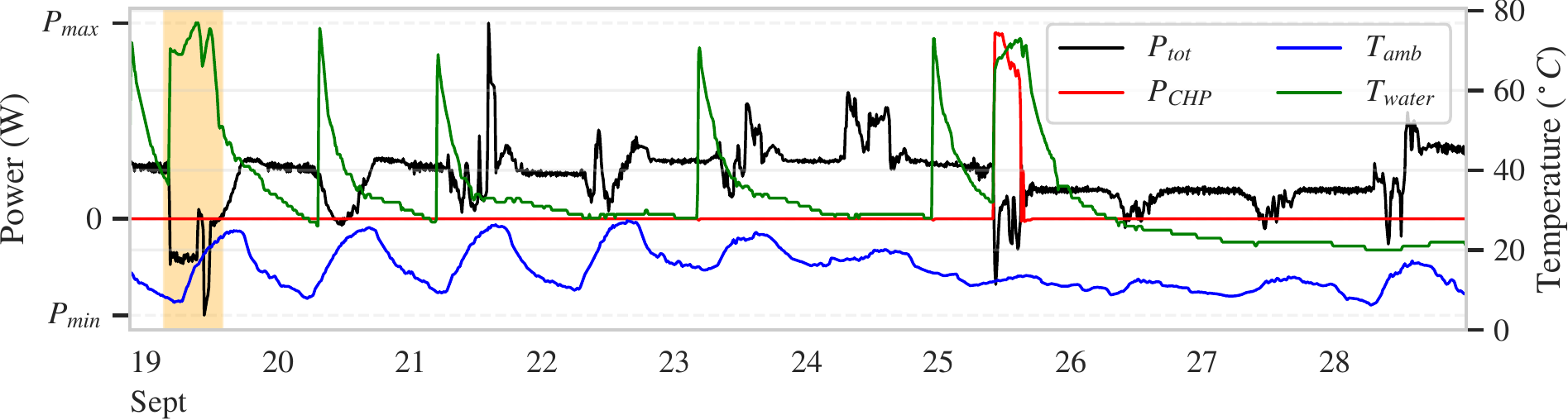}
\caption{Sensory data from the CHP and total electrical power. Upper: normal operation. Lower: an example of $P_{CHP}$ sensor malfunction is highlighted in yellow.}
\label{fig:CHP_data}
\end{figure}

The CHP is a complex industrial machinery which burns natural gas in order to produce heat and electrical power.
It is controlled by an algorithm where only some aspects are known, so its behavior is mostly unclear and not well predictable.  
It is known that important control signals are the ambient outside temperature $T_{amb}$, the  internal water temperature $T_{water}$, the generated electrical power output $P_{CHP}$, and the total electricity demand of the facility $P_{tot}$.
Fig.~\ref{fig:CHP_data} shows examples of recorded data from the CHP as well as  $P_{tot}$ of the facility.
 $T_{amb}$ has a known strong influence as the CHP is off in the summer period and more or less continuously on in winter and cold periods.
In the transition seasons (spring and fall), the CHP sometimes turns on (night of 25$^\m{th}$ Sept.), sometimes just heats up its internal water (nights of 20$^\m{st}$, 21$^\m{st}$, and 23$^\m{rd}$ Sept.), or  exhibits a fast switching behavior (e.g. 20$^\m{st}$, and 27$^\m{th}$ Oct.).
Even though the CHP usually turns on for a couple of hours (e.g.\ 25$^\m{th}$ Sept.) at rare instances it just turns on for a very short time (e.g.\ 28$^\m{th}$ Oct.).

Due to the complicated operational pattern, it is already hard to make a detector for the CHP operational state even with full access to the measurement data. 
Additionally, the sensors measuring the CHP output power are prone to failure which can be observed in the yellow highlighted area in the bottom side of Fig.~\ref{fig:CHP_data}. 
During that 10-hour period, the CHP did produce electrical power, even though the sensor reading does not indicate this. 
The total electrical power drawn from the grid, $P_{tot}$, provides a much more stable measurement signal. 
The signature of the CHP is clearly visible in the total power signal and we propose to estimate  $P_{CHP}$ from this signal,  but many variables also affects the total load, e.g.\ PV system, changing workloads, etc.

We focus on estimating the power output but where the estimate power value should only be accurate within a certain range. 
Thus,  the problem is formulated as time-series classification instead of regression, where each class represents a certain range of output values.
The class labels are calculated directly from the $P_{CHP}$ sensor measurement as the mean power output of a fixed-length sliding window.
Due to frequent sensor malfunctioning, as displayed in Fig.~\ref{fig:CHP_data}, the resulting classification problem is subject to label noise.


\section{\acrfull{method}}\label{sec:method}
\subsubsection{Architecture and loss function} 
\begin{figure}[tb]
    \centering
    \includegraphics[width=0.55\linewidth]{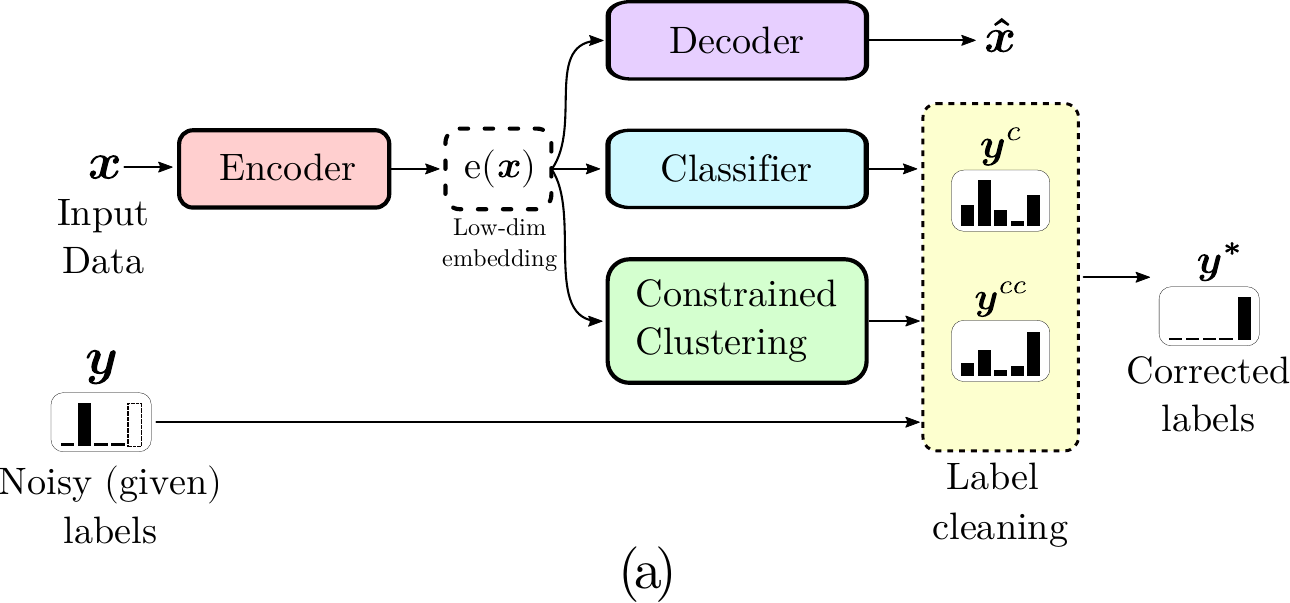}\:\:
    \includegraphics[width=0.4\linewidth]{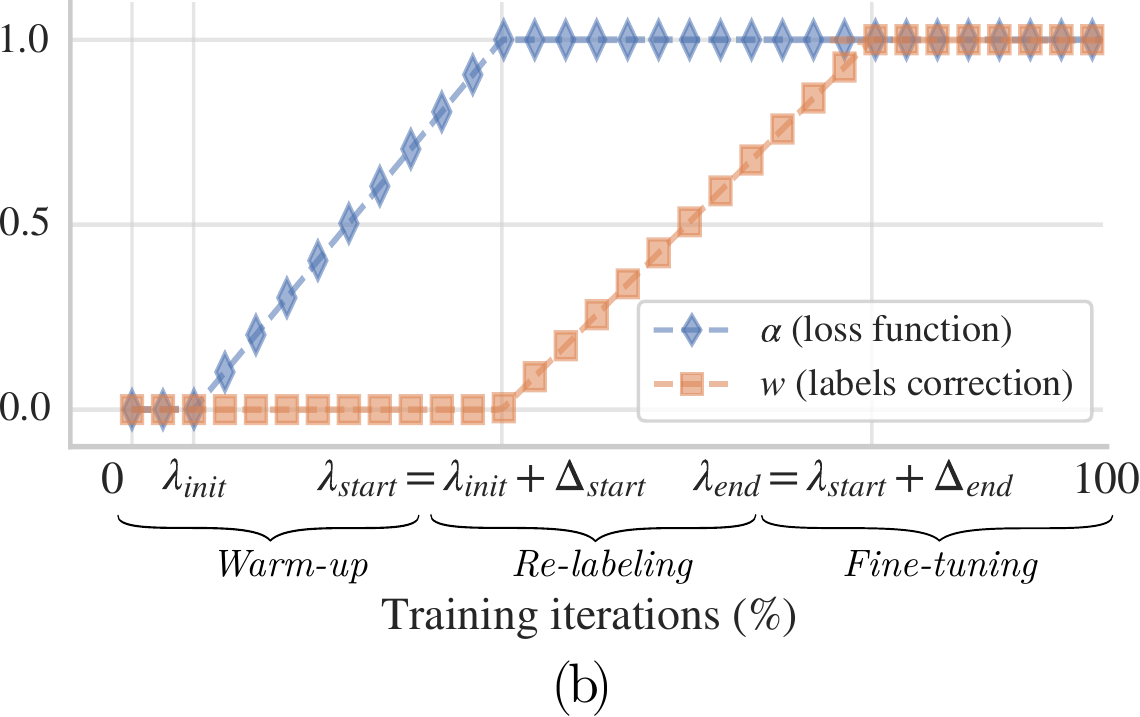}
    \caption{\acrshort{method} processing architecture (a) and the dynamics of the parameters $\alpha$ and $w$ during the training epochs (b). }
    \label{fig:CNN}
\end{figure}

In this work, column vectors are denoted in bold (e.g. $\bm{x}$). As described in Sec.~\ref{sec:CHP}, we treat the challenge to model time series data as
a classification problem to predict  averaged characteristics of the process using windowing techniques.
Hence, we deal with a supervised \textit{k}-class classification problem setting with a dataset of $n$ training examples $\mathcal{D}= \{ (\bm{x}_i, \bm{y}_i), i=1,...,n\}$ with $\bm{y}_i \in \{ 0, 1 \} ^k $  being the one-hot encoding label for sample $\bm{x}_i$. Thereby, {\bf label noise} is present, i.e.\ we expect that $\bm{y}_i$ is wrong for a substantial (and possibly skewed) amount of instances in the training set.

The overall processing architecture of the proposed approach is shown in Fig.~\ref{fig:CNN}(a).
The \textit{autoencoder} ($f_{ae}$), represented by the \textit{encoder} ($e$) and the \textit{decoder}, provides a strong surrogate supervisory signal for feature learning \cite{jawed2020self} that is not affected by the label noise. 
Two additional components, a \textit{classifier} network ($f_c$) and a \textit{constrained clustering} module ($f_{cc}$) are introduced, which independently propose class labels as output. 
Each of the three processing pipelines share the same embedding representation, created by the encoder,  and each output module is associated with one separate contribution to the total loss function.

For the autoencoder, a typical reconstruction loss is utilized:
\begin{equation}
    \mathcal{L}_{ae} = \frac{1}{n} \sum_{i = 1}^{n} \left ( \bm{\hat{x}}_i - \bm{x}_i \right ) ^2 ,
\end{equation}
where $\bm{\hat{x}}_i$ is the output of the autoencoder given the input $\bm{x}_i$.

Cross entropy is used as loss function for the classification network output, 
\begin{equation}\label{eq:CE}
    \mathcal{L}_{c} = -\frac{1}{n} \sum_{i=1}^{n}  \bm{y}_{i}^T\cdot \log (\bm{p}^c_{i}),
\end{equation}
where $\bm{p}^c_i$ are the \textit{k}-class softmax probabilities produced by the model for the training sample $i$, i.e. $\bm{p}^c_i = \text{softmax}(f_c(\bm{x}_i))$.

For the constraint clustering loss, we first initialize the cluster center $\bm{C} \in \mathbb{R}^{d \times k}$ in the \textit{d}-dimensional embedding space, with the $k$-means clustering of the training samples.
Then, inspired by the good results achieved in \cite{zeghidour2020wavesplit}, we constrain the embedding space to have small intra-class and large inter-class distances, by iteratively adapting $\bm{C}$. 
The resulting clustering loss is given by:
\begin{equation}
    \mathcal{L}_{cc} = \frac{1}{n}\sum_{i=1}^{n} \Bigg[ {\underbrace{  \left \|\bm  e(\bm{x}_i) - \bm{C}_{\bm{y}_i} \right \| ^ 2 _2}_{\text{intra-class}}} + {\underbrace{\log \sum_{j=1}^{k} \exp \left ( - \left \|\bm e(\bm{x}_i) - \bm{C}_j \right \| _2 \right ) }_{\text{inter-class}}} \Bigg]  + \ell_{reg}.
\end{equation}
The entropy regularization $\ell_{reg} = - \sum_i^k \min_{i \neq j} \log \left \| \bm{C}_i - \bm{C}_j \right \|_2$ is aimed to create well separated embedding for different classes  \cite{sablayrolles2019spreading}.

The final total loss function is given by the sum of those contributions, 
\begin{equation}\label{eq:L}
    \mathcal{L} =   \mathcal{L}_{ae}  + \alpha \left( \mathcal{L}_{c} +  \mathcal{L}_{cc} +  \mathcal{L}_\rho \right),
\end{equation}
where we introduced the dynamic parameter $0\leq \alpha \leq 1$ which changes during the  training  (explained below).
We also add the regularization loss $\mathcal{L}_\rho$ to prevent the assignment of all labels to a single class, 
$   \mathcal{L}_\rho = \sum_{j=1}^k \bm{h}_j\cdot \log \frac{\bm{h}_j}{\bm{p}^\rho_{j}}
$,
where $\bm{h}_j$ denotes the prior probability distribution for class \textit{j}, which assume to be uniformly distributed to $1/k$. The term $\bm{p}^\rho_{j}$ is the mean softmax probability of the model for class \textit{j} across all samples in dataset which we  approximate using mini-batches as done in previous works \cite{arazo2019unsupervised}.

\subsubsection{Re-Labeling strategy} We do not compute the total loss function in Eq.~\eqref{eq:L} with the given (noisy) label $\bm y_i$. 
But, we estimate the true label for each data sample $\bm{y^*}_i$ by taking the weighted average of the given training label $\bm y_i$, and the pseudo-labels proposed by the classifier $\bm y^{c}_i$ and the constraint clustering $\bm y^{cc}_i$. 

In order to increase robustness of the labels proposed by the classifier ($\bm y^{c}_i$), we take for each data sample the exponentially averaged probabilities during the last five training epochs, with the weighting factor $\tau_t \sim e^{\frac{t-5}{2}}$:
\begin{equation}\label{eq:C}
    \bm y^{c}_{i}=  \sum_{\m{last\:5\:epochs\: } t} \tau_t \; [\bm p_i^c]_t.
\end{equation}

The label from constraint clustering ($\bm{y}^{cc}_{i}$) is determined by the distances of the samples to the cluster centers in the embedding space:
\begin{equation}
    \label{eq:CC}
    \bm y^{cc}_{i} = \text{softmin}_j \big(\| \bm e(\bm{x}_i) -\bm C_j \|_2\big)
\end{equation}

Then, the corrected label $\bm{y^*}_i$ (in one hot-encoding) is produced by selecting the class corresponding to the maximum entry:    
 \begin{equation}\label{eq:y*}
     \bm y^*_{i} = \text{argmax} \big[(1-w) \,\bm y_{i}+ w\,(\bm y^{c}_{i} +\bm y^{cc}_{i}) \big]
\end{equation}
where the dynamic weighting factor $0\leq w \leq 1$ is function of the training epoch $t$ and to be discussed below when explaining the training dynamics.

\subsubsection{Training Dynamics} A key aspect of our proposed approach is to dynamically change the loss function, as well as the label correction mechanism, during the training. 
This is achieved by changing the parameters $\alpha$ (loss function) and $w$ (label correction mechanism) as depicted in Fig.~\ref{fig:CNN}(b). The training dynamics is completely defined by the three hyper-parameters $\lambda_{init}$, $\Delta_{start}$ and $\Delta_{end}$.

Initially, we start with $\alpha=0$ and  $w=0$, and  only train the autoencoder from epoch $t=0$ to epoch $t=\lambda_{init}$. 
At the training epoch $t=\lambda_{init}$, $\alpha$ is ramped up linearly until it reaches $\alpha=1$ in training epoch $t=\lambda_{start}=\lambda_{init}+\Delta_{start}$. 
The purpose of this first \textit{warm-up} period is an unsupervised initialization of the embedding space with slowly turning on the supervision of the given labels.
The dominant structure of the clean labels is learned, as neural networks tend to learn the true labels, rather than overfit to the noisy ones, at early training stages \cite{zhang2016understanding,arpit2017closer}.
Then, we also increase $w$ linearly from zero to one, between epochs $t=\lambda_{start}$ to $t=\lambda_{start}+\Delta_{end}=\lambda_{end}$, thereby turning on the label correction mechanism (\textit{re-labeling}).
After training epoch $t=\lambda_{end}$ until the rest of the training, we keep $\alpha=1$ and $w=1$ which means we are fully self-supervised  where the given training labels do not enter directly anymore (\textit{fine-tuning}). 
We summarize the \acrshort{method} and display the pseudo-code in Algorithm~\ref{alg:method}.

\begin{algorithm}[t]
\DontPrintSemicolon
\footnotesize
  \KwInput{Data $\{ (\bm{x}_i, \bm{y}_i)\}_n$, autoencoder $f_{ae}$, classifier $f_c$, constraint clustering $f_{cc}$,  hyper-parameters: $\lambda_{init}$,   $\Delta_{start}$,  $ \Delta_{end}$.
  }
  Init hyper-parameter ramp-up functions $w_t$ and $\alpha_t$ \tcp*{see Fig.~\ref{fig:CNN}(b)}
  \For{training epoch $t=0$ \KwTo $t_{end}$ }{
  Fetch mini-batch data $\{ (\bm{x}_i, \bm{y}_i)\}_b$ at current epoch $t$\\
  \For{$i=1$ \KwTo $b$}{
  \If{$t==\lambda_{init}$}{
        $f_{cc}\gets$ $k$-means$(\bm{e}(\bm{x}_i))$ \tcp*{Initialize constraint clustering}
        }
    $\bm{\hat{x}}_i = f_{ae}(\bm{x}_i)$ \tcp*{Auto-encoder forward pass}
    $\bm y^c_i \gets$ Eq.\eqref{eq:C} \tcp*{Classifier forward pass}
    $\bm{y}^{cc}_i\gets$ Eq.\eqref{eq:CC}\tcp*{Constraint clustering output}
    Adjust $w\gets w_t$,  $\alpha\gets \alpha_t$  \tcp*{Label correction and loss parameters}
    $\bm{y^*}_i \gets$ Eq.\eqref{eq:y*} \tcp*{Re-labeling}
    $\mathcal{L}\gets$ Eq.\eqref{eq:L}  \tcp*{Evaluate loss function}
    Update $f_{ae}, f_c, f_{cc}$ by SGD on $\mathcal{L}$
    }
    }
\caption{\acrshort{method}: \acrlong{method}.}\label{alg:method}
\end{algorithm}


\section{Experimental setup}

\subsubsection{Label Noise}\label{sec:noise}
True labels are corrupted by a \textit{label transition matrix} T \cite{song2020learning}, where $T_{ij}$ 
is the probability of the label $i$ being flipped into label $j$. 
For all the experiments, we corrupt the labels with \textit{symmetric} (unstructured) and \textit{asymmetric} (structured) noise with noise ratio $\epsilon \in [0, 1]$.
For symmetric noise, a true label is randomly assigned to other labels with equal probability, i.e.  $T_{ii} = 1 - \epsilon$ and $ T_{ij}= \frac{\epsilon}{k-1}$ ($i\neq j$), with $k$ the number of classes.
For asymmetric noise,  a true label is mislabelled by shifting it by one, i.e.\ $T_{ii} = 1 - \epsilon$ and $T_{(j+1\m{mod} k )j} = \epsilon$ ($i\neq j$).
For the estimation of the CHP power, we also analyze another kind of structured noise which we call \textit{flip} noise, where a true label is only flipped to zero, i.e.\  $ T_{ii} = 1 - \epsilon$ and $T_{i0} = \epsilon$. 
This mimics 
sensor failures, where a broken sensor produces a constant output regardless of the real value.
Note that learning with structured noise is much harder than with unstructured noise \cite{frenay2013classification}.

\subsubsection{Network architecture}
Since \acrshort{method} is model agnostic, we use CNNs in the experiments, as  these are currently the SotA deep learning network topology for time-series classification \cite{wang2017time,fawaz2019deep}.
The encoder and decoder have a symmetric structure with 4 convolutional blocks.
Each block is composed by a 1D-conv layer followed by batch normalization \cite{ioffe2015batch}, a ReLU activation and a dropout layer with probability 0.2.
The dimension of the shared embedding space is 32.
For the classifier we use a fully connected network with 128 hidden units and \textit{\#classes} outputs.
We use the Adam optimizer \cite{kingma2014adam} with an initial learning rate of 0.01 for 100 epochs.
Such high value for the initial learning rate helps to avoid overfitting of noisy data in the early stages of training \cite{zhang2016understanding,arpit2017closer}.
We halve the learning rate every $20\%$ of training (20 epochs).
In the experiments, we assume to not have access to any clean data, thus it is not possible to use a validation set, and the models are trained without early stopping.
The \acrshort{method} hyper-parameters $\lambda_{init} = 0$, $\Delta_{start} = 25$ and $\Delta_{end} = 30$ are used if not specified otherwise.
Further implementation details are reported in the supplementary material~\cite{castellaniSuppl2021}, including references to the availability of code and data sets.

\subsubsection{Comparative methods}\label{sec:methods}
In order to make a fair comparison, we use the same neural network topology throughout all the experiments.
A baseline method, which does not take in accout any label noise correction criteria, is a CNN classifier \cite{wang2017time} trained with
cross-entropy loss function of Eq.~\eqref{eq:CE}, which we refer to as \textit{CE}. 
We compare to \textit{MixUp} \cite{zhang2017mixup}, which is a data augmentation technique that exhibits strong robustness to label noise.
In \textit{MixUp-BMM} \cite{arazo2019unsupervised} a two-component beta-mixture model is fitted to the loss distribution and training with bootstrapping loss is implemented.
\textit{SIGUA} \cite{han2020sigua} implements stochastic gradient ascent on likely mislabeled data, thereby trying to reduce the effect of noisy labels.
Finally, in \textit{Co-teaching} \cite{sugiyama2018co} two networks are simultaneously trained which inform each other about which training examples to keep.
The algorithms \textit{SIGUA} and \textit{Co-teaching} assume that the noise level $\epsilon$ is known. 
In our experiments, we use the true value of $\epsilon$ for those approaches, in order to create an upper-bound of their performance.
All the hyper-parameters of the investigated algorithm are set to their default and recommended values.

\subsubsection{Implementation details}
For the problem of estimating CHP power output, the raw data consists of 78 days of measurement with a sampling rate of 1 sample/minute.
As preprocessing, we do a re-sampling to 6 samples/hour.
The CHP should have a minimal on-time of one hour, in order to avoid too rapid switching which would damage the machine. 
However, during normal operation, the CHP is controlled in a way that on- and off-time periods are  around 4 to 8 hours. 
Due to these time-scales, we are interested in the power output on a scale of 6 hours, which means we use a sliding window with a size of 6 hours (36 samples) and a stride of 10 minutes (1 sample).
Therefore, the preprocessing of the three input variables  $P_{tot}$, $T_{water}$, and  $T_{amb}$ lead to $\mathbb{R}^{(36 \times 3)}$-dimensional data samples.
For generating the labels, we use 5 power output levels, linearly spaced from 0 to $P_{CHP,max}$, and correspondingly to a five-dimensional one-hot encoded label vector $\bm{y}_i\in \mathbb{R}^5$.
For every dataset investigated, we normalize the dataset to have zero mean and unit standard deviation, and we randomly divide the total available data in train-set and test-set with a ratio 80:20.

\subsubsection{Evaluation measures}
 
To evaluate the performance, we report the \textit{averaged $\mathcal{F}_1$-score} on the test-set, where the well-known $F_1$-scores are calculated for each class separately and then averaged via arithmetic mean, 
$\mathcal{F}_1 = \frac{1}{k} \sum_{j=1}^{k} F_{1,j} $.
This formulation of the $F_1$-score results in a larger penalization when the model do not perform well in the minority class, in cases with class imbalance. 
Other metrics, such as the accuracy, show qualitatively similar results and are therefore not reported in this manuscript.
In order to get performance statistics of the methods,
all the experiments have been repeated 10 times with different random initialization. The non-parametric statistical  \textit{Mann-Whitney  U  Test} \cite{mcknight2010mann}
is used to compare the \acrshort{method} against the SotA algorithms.

\section{Results and discussion}

\subsubsection{Benchmarks Datasets}
We evaluate the proposed \acrshort{method} on publicly available time-series classification datasets from UCR repository \cite{UCRArchive2018}.
We randomly choose 10 datasets with different size, length, number of classes and dimensions in order to try to avoid bias in the data.
A summary of the datasets is given in Table~\ref{tab:UCR_data}. 
\begin{wraptable}[14]{r}{0.5\textwidth}
    \centering
    \scriptsize
    \caption{UCR Single-variate and Multi-variate dataset description.}
    \label{tab:UCR_data}
    \begin{tabular}{l l l l l}
    \toprule
        \textbf{Dataset} & \textbf{Size} & \textbf{Length} & \textbf{\#classes} & \textbf{\#dim}  \\
    \midrule
         ArrowHead$^\dagger$ & 211 & 251 & 3 & 1 \\
         CBF & 930 & 128 & 3 & 1 \\
         Epilepsy & 275 & 206 & 4 & 3 \\
         FaceFour$^\dagger$ & 112 & 350 & 4 & 1 \\
         MelbourneP. & 3633 & 24 & 10 & 1 \\
         NATOPS & 360 & 51 & 6 & 24 \\
         OSULeaf$^\dagger$ & 442 & 427 & 6 & 1 \\
         Plane & 210 & 144 & 7 & 1 \\
         Symbols$^\dagger$ & 1020 & 398 & 6 & 1 \\
         Trace$^\dagger$ &  200 & 275 & 4 & 1\\
    \bottomrule
    \multicolumn{5}{l}{$\dagger$ results reported in the supplementary material~\cite{castellaniSuppl2021}.}
    \end{tabular}
\end{wraptable}

We show a representative selection of all comparisons of the results in Table~\ref{tab:ucr}. 
Without any label noise, CE is expected to provide very good results. We observe that our \acrshort{method} gives similar or better $\F_1$ scores than the CE method on 9 out of 10 datasets without label noise.
Considering all algorithms and datasets, with symmetric noise we achieve statistically significantly better scores in 62, similar scores in 105, and worse scores in 33 experiments out of a total of 200 experiments\footnote{Number of experiments: 10 datasets $\times$ 4 noise levels $\times$ 5 algorithms = 200. Each experiment consists of 10 independent runs. }. 
For the more challenging case of asymmetric noise, the \acrshort{method}-results are 86 times significantly better, 97 times equal, and 17 times worse than SotA algorithms.

\begin{table*}[th!]
    \centering
    \scriptsize
    \caption{$\F_1$ test scores on UCR datasets.
    The best results per noise level are underlined. In parenthesis the results of a Mann–Whitney U test with $\alpha=0.05$ of \acrshort{method} against the other approaches: SREA $\F_1$ is significantly higher ($+$), lower ($-$) or not significant ($\approx$).}
    \label{tab:ucr}
    \setlength{\tabcolsep}{1pt}
    \renewcommand{\arraystretch}{0.5}
    \begin{tabular*}{\linewidth}{l l @{\extracolsep{\fill}} l l l l l l l}
    \toprule
    \textbf{Dataset}&\textbf{Noise}&\textbf{\%}&\textbf{CE}&\textbf{MixUp}&\textbf{M-BMM}&\textbf{SIGUA}&\textbf{Co-teach}&\textbf{\acrshort{method}}\\
    \midrule

    \multirow{8}{0.13\linewidth}{\textit{CBF}}
    & - & 0     & 1.000 $(+)$ & 0.970 $(+)$ & 0.886 $(+)$ & 1.000 $(+)$ & 0.997 $(+)$ & \underline{1.000}\\
    \cmidrule{2-9}
    & \multirow{2}{*}{Symm}& 15  & 0.943 $(+)$ & 0.923 $(+)$ & 0.941 $(+)$ & 0.976 $(+)$ & 0.923 $(+)$ & \underline{1.000} \\
    & & 30  & 0.780 $(+)$ & 0.799 $(+)$ & 0.932 $(+)$ & 0.923 $(+)$ & 0.833 $(+)$ & \underline{0.998} \\
    \cmidrule{2-9}
    & \multirow{2}{*}{Asymm} & 10  & 0.973 $(+)$ & 0.956 $(+)$ & 0.920 $(+)$ & 0.989 $(+)$ & 0.963 $(+)$ & \underline{1.000} \\
    & & 20  & 0.905 $(+)$ & 0.897 $(+)$ & 0.949 $(+)$ & 0.980 $(+)$ & 0.900 $(+)$ & \underline{1.000} \\
    \midrule
    
    \multirow{8}{0.13\linewidth}{\textit{Epilepsy}}
    & - & 0 & \underline{0.974} $(\approx)$ & 0.955 $(+)$ & 0.926 $(+)$ & \underline{0.978} $(\approx)$ & 0.971 $(+)$ & \underline{0.973} \\
    \cmidrule{2-9}
    & \multirow{2}{*}{Symm}& 15 &  \underline{0.890} $(\approx)$ &  \underline{0.913} $(\approx)$ &  \underline{0.899} $(\approx)$ &  \underline{0.884} $(\approx)$ &  \underline{0.861} $(\approx)$ &  \underline{0.861}\\
    & & 30 & 0.784 $(-)$ & 0.823 $(-)$ & 0.805 $(-)$ &  \underline{0.741} $(\approx)$ &  \underline{0.744} $(\approx)$ &  \underline{0.708} \\
    \cmidrule{2-9}
    & \multirow{2}{*}{Asymm} & 10 & \underline{0.919} $(\approx)$ & \underline{0.930} $(\approx)$ & \underline{0.847} $(\approx)$ & \underline{0.905} $(\approx)$ & \underline{0.919} $(\approx)$ & \underline{0.888}\\
    & & 20 & \underline{0.861} $(\approx)$ & 0.894 $(-)$ & 0.891 $(-)$ & \underline{0.826} $(\approx)$ & \underline{0.863} $(\approx)$ & \underline{0.825} \\
    \midrule

    \multirow{8}{0.13\linewidth}{\textit{Melbourne}}
    & - & 0 & \underline{0.923} $(\approx)$ & 0.879 $(+)$ & 0.773 $(+)$ & \underline{0.918} $(\approx)$ & \underline{0.913} $(\approx)$ & \underline{0.911}\\
    \cmidrule{2-9}
    & \multirow{2}{*}{Symm}& 15 & 0.869 $(+)$ & 0.870 $(+)$ & 0.856 $(+)$ & \underline{0.883} $(\approx)$ & \underline{0.886} $(\approx)$ & \underline{0.883}\\
    & & 30 & 0.826 $(+)$ & \underline{0.858} $(\approx)$ & \underline{0.870} $(\approx)$ & \underline{0.855} $(\approx)$ & 0.876 $(-)$ & \underline{0.862}\\
    \cmidrule{2-9}
    & \multirow{2}{*}{Asymm} & 10 & 0.898 $(+)$ & 0.877 $(+)$ & 0.860 $(+)$ & 0.899 $(+)$ & 0.897 $(+)$ & \underline{0.911}\\
    & & 20 & 0.865 $(+)$ & 0.861 $(+)$ & 0.851 $(+)$ & 0.858 $(+)$ & \underline{0.893} $(\approx)$ & \underline{0.903}\\
    \midrule
    
    \multirow{8}{0.13\linewidth}{\textit{NATOPS}}
    & - & 0 & \underline{0.858} $(\approx)$ & \underline{0.801} $(\approx)$ & 0.711 $(+)$ & \underline{0.848} $(\approx)$ & \underline{0.835} $(\approx)$ & \underline{0.866}\\
    \cmidrule{2-9}
    & \multirow{2}{*}{Symm}& 15 & \underline{0.779} $(\approx)$ & 0.718 $(+)$ & 0.702 $(+)$ & \underline{0.754} $(\approx)$ & \underline{0.761} $(\approx)$ & \underline{0.796}\\
    & & 30 & \underline{0.587} $(\approx)$ & 0.580 $(+)$ & 0.602 $(+)$ & 0.593 $(+)$ & 0.673 $(+)$ & \underline{0.670}\\
    \cmidrule{2-9}
    & \multirow{2}{*}{Asymm} & 10 & 0.798 $(+)$ & \underline{0.822} $(\approx)$ & 0.756 $(+)$ & 0.764 $(+)$ & 0.790 $(+)$ & \underline{0.829}\\
    & & 20 & \underline{0.703} $(\approx)$ & \underline{0.763} $(\approx)$ & \underline{0.762} $(\approx)$ & \underline{0.698} $(\approx)$ & \underline{0.733} $(\approx)$ & \underline{0.762}\\
    \midrule
    
    \multirow{8}{0.13\linewidth}{\textit{Plane}}
    & - & 0 & \underline{0.995} $(\approx)$ & 0.962 $(+)$ & 0.577 $(+)$ & 0.981 $(+)$ & \underline{0.990} $(\approx)$ & \underline{0.998}\\
    \cmidrule{2-9}
    & \multirow{2}{*}{Symm}& 15 & 0.930 $(+)$ & 0.953 $(+)$ & 0.873 $(+)$ & \underline{0.971} $(\approx)$ & \underline{0.981} $(\approx)$ & \underline{0.983}\\
    & & 30 & 0.887 $(+)$ & 0.902 $(+)$ & \underline{0.943} $(\approx)$ & 0.862 $(+)$ & \underline{0.941} $(\approx)$ & \underline{0.944}\\
    \cmidrule{2-9}
    & \multirow{2}{*}{Asymm} & 10 & \underline{0.981} $(\approx)$ & \underline{0.986} $(\approx)$ & 0.648 $(+)$ & \underline{0.986} $(\approx)$ & \underline{0.990} $(\approx)$ & \underline{0.976}\\
    & & 20 & \underline{0.952} $(\approx)$ & \underline{0.923} $(\approx)$ & 0.751 $(+)$ & \underline{0.976} $(\approx)$ & \underline{0.990} $(\approx)$ & \underline{0.966}\\
    \bottomrule
    
    \end{tabular*}

\end{table*}

\subsubsection{CHP power estimation}
\begin{table*}[tbh]
    \centering
    \scriptsize
    \setlength{\tabcolsep}{3pt}
    \caption{ $\F_1$ test scores of the CHP power estimation. Same notation as Table~\ref{tab:ucr}.}
    \label{tab:CHP_results}
    \begin{tabular*}{\textwidth}{ l l @{\extracolsep{\fill}}  l l l l l l}
    \toprule
    \textbf{Noise}&\textbf{\%}&\textbf{CE}&\textbf{MixUp}&\textbf{M-BMM}&\textbf{SIGUA}&\textbf{Co-teach}&\textbf{\acrshort{method}}\\
    \midrule
    \textbf{-} & 0 & \underline{0.980} $(\approx)$ & 0.957 $(+)$ & 0.882 $(+)$ & \underline{0.979} $(\approx)$ & \underline{0.974} $(\approx)$ & \underline{0.979} \\
    \midrule
    \multirow{4}{*}{\textbf{\textit{Symmetric}}} & 15 & 0.931 $(+)$ & 0.934 $(+)$ & 0.903 $(+)$ & \underline{0.954} $(\approx)$ & \underline{0.950} $(\approx)$ & \underline{0.960}\\
    & 30 & 0.856 $(+)$ & 0.910 $(+)$ & 0.897 $(+)$ & 0.912 $(+)$ & 0.920 $(+)$ & \underline{0.938}\\
    & 45 & 0.763 $(+)$ & 0.883 $(+)$ & 0.895 $(+)$ & 0.867 $(+)$ & 0.886 $(+)$ & \underline{0.918}\\
    & 60 & 0.661 $(+)$ & \underline{0.761} $(\approx)$ & 0.692 $(+)$ & \underline{0.817} $(\approx)$ & \underline{0.839} $(\approx)$ & \underline{0.800}\\
    \midrule
    \multirow{4}{*}{\textbf{\textit{Asymmetric}}} & 10 & \underline{0.954} $(\approx)$ & 0.945 $(+)$ & 0.893 $(+)$ & \underline{0.959} $(\approx)$ & \underline{0.964} $(\approx)$ & \underline{0.961}\\
    & 20 & 0.924 $(+)$ & 0.925 $(+)$ & 0.899 $(+)$ & \underline{0.935} $(\approx)$ & \underline{0.938} $(\approx)$ & \underline{0.946}\\
    & 30 & \underline{0.895} $(\approx)$ & \underline{0.909} $(\approx)$ & 0.873 $(+)$ & \underline{0.916} $(\approx)$ & \underline{0.923} $(\approx)$ & \underline{0.919}\\
    & 40 & 0.807 $(-)$ & 0.848 $(-)$ & 0.784 $(-)$ & 0.836 $(-)$ & \underline{0.876} $(-)$ & 0.287\\
    \midrule
    \multirow{4}{*}{\textbf{\textit{Flip}}} & 10 & \underline{0.970} $(\approx)$ & 0.950 $(+)$ & 0.860 $(+)$ & 0.965 $(+)$ & \underline{0.973} $(\approx)$ & \underline{0.971}\\
    & 20 & 0.942 $(+)$ & 0.942 $(+)$ & 0.860 $(+)$ & 0.945 $(+)$ & \underline{0.962} $(\approx)$ & \underline{0.963}\\
    & 30 & 0.908 $(+)$ & 0.919 $(+)$ & 0.880 $(+)$ & 0.923 $(+)$ & 0.792 $(+)$ & \underline{0.956}\\
    & 40 & 0.868 $(+)$ & 0.791 $(+)$ & 0.696 $(+)$ & 0.779 $(+)$ & 0.623 $(+)$ & \underline{0.945}\\
    \bottomrule
    \end{tabular*}
\end{table*}

\begin{figure}[tb]
    \centering
        \includegraphics[width=0.35\textwidth]{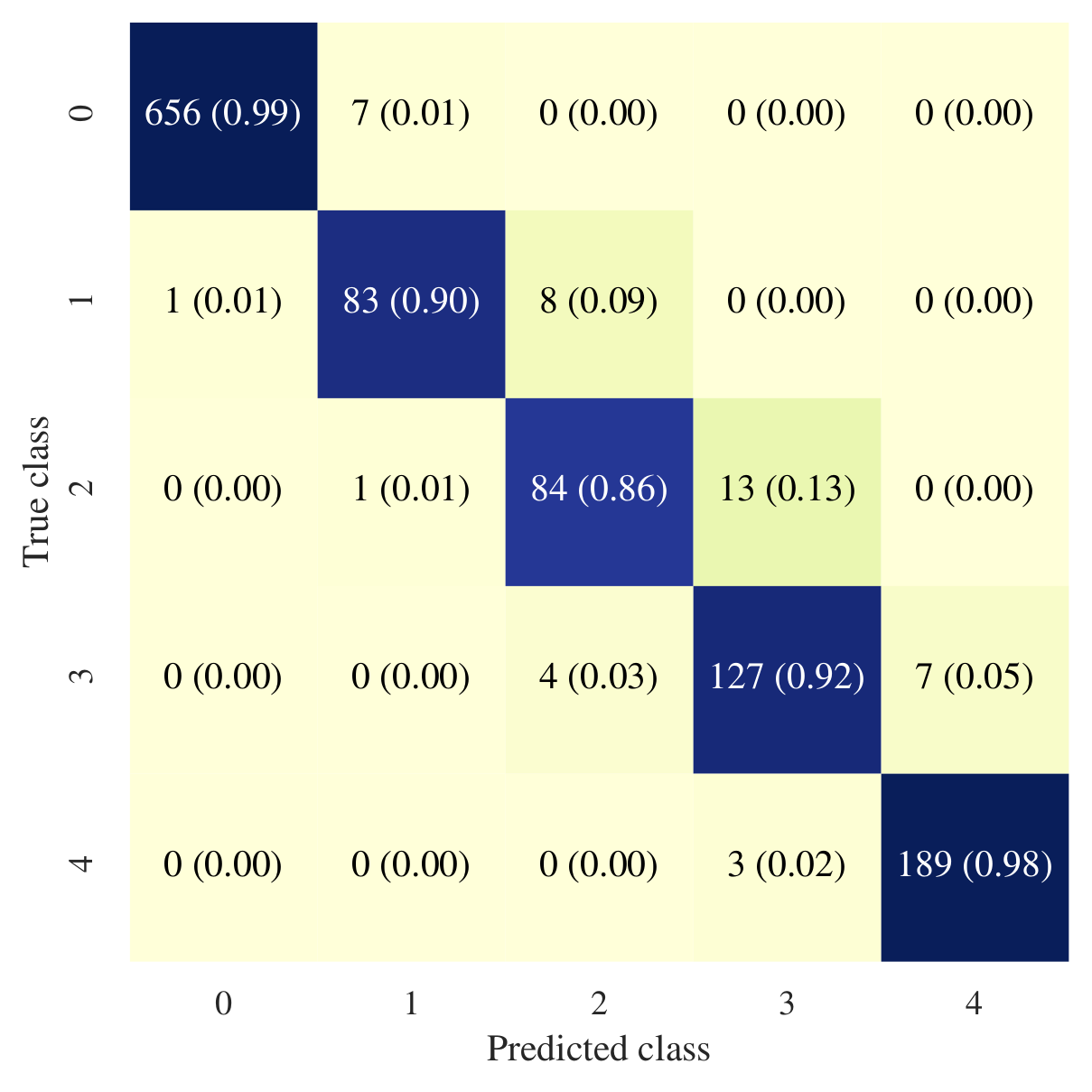}\:
       \centering
        \includegraphics[width=0.35\textwidth]{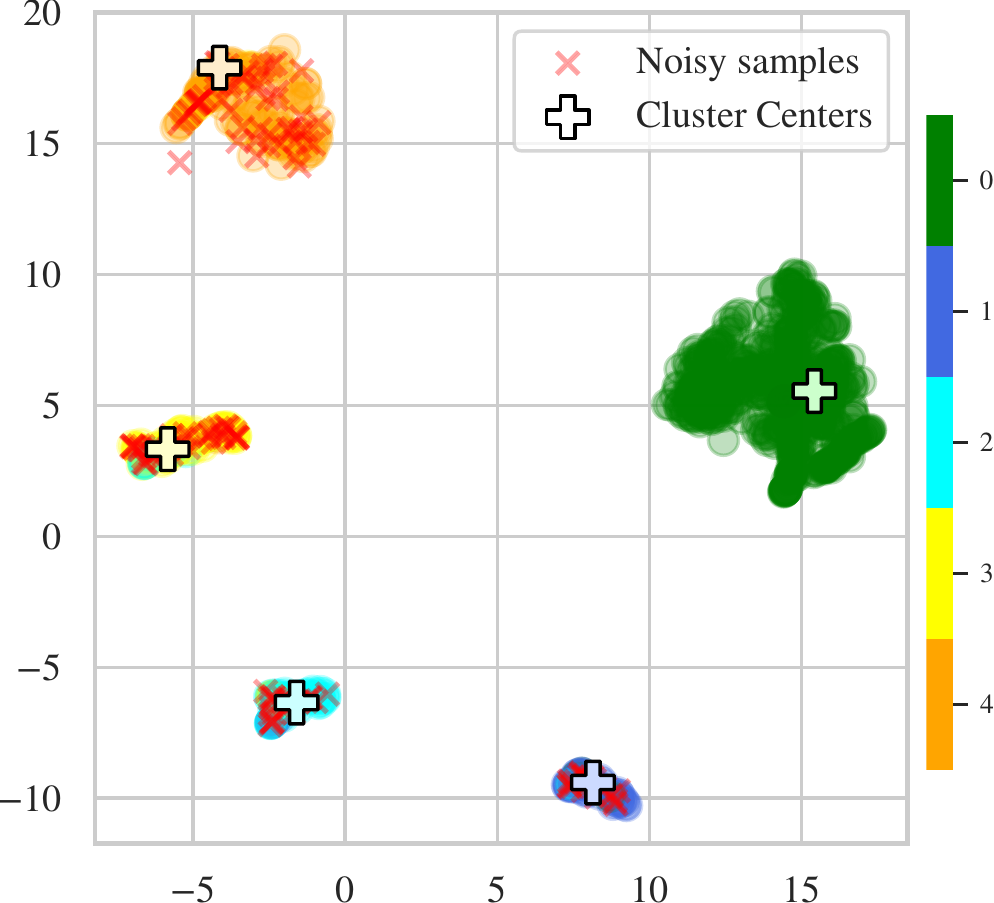}
    \caption{Confusion matrix of the corrected labels (left) and embedding space of the train-set (right) of the CHP power estimation, corrupted with 30\% flip noise.}
    \label{fig:cm_umap}
\end{figure}

Table~\ref{tab:CHP_results} shows the results of the estimation of the CHP output power level. 
Without labels noise, the proposed approach has comparable performance to CE, with an average  $\F_1$-score of 0.979. 
This implies that we successfully solve the  CHP power estimating  problem  by analyzing the total load with an error rate less than 2\%.
When the training labels are corrupted with low level of symmetric label noise, $\epsilon \leq 0.4$, \acrshort{method} consistently outperforms the other algorithms. With higher level of symmetric noise we achieve a comparable performance to the other algorithms.
Under the presence of asymmetric label noise, \acrshort{method} shows  a performance comparable  to  the other SotA algorithms.
Only for a  highly unrealistic asymmetric noise level of 40\%, the performance is significantly worse than the SotA. This indicates that, during the warm-up and relabelling phase, the network is not able to learn the true labels but also learns the wrong labels induced by the structured noise. During the fine-tuning phase, the feedback of the wrongly labeled instances is amplified and the wrong labels a reinforced. 
For flip noise, which reflects sensors failures, \acrshort{method} retains a high performance and outperforms all other SotA algorithms up to noise levels of 40\%.
For low noise levels up to 20\%, \acrshort{method} has similar performance to  Co-teaching, but without the need to know the amount of noise. 

In Fig.~\ref{fig:cm_umap} (left) we show the label confusion matrix (with in-class percentage in parentheses) of \acrshort{method} for the resulting corrected labels for the case of 30\% of flip label noise.  
The corrected labels have 99\% and 98\% accuracy for the fully off- (0) and on-state (4), respectively. The intermediate power values' accuracies are 90\% for state 1, 86\% for state 2 and 92\% for state 3.
As an example, a visualization of the 32 dimensional embedding using the  UMAP \cite{McInnes2018UMAPUM} dimension reduction technique is shown in Fig.~\ref{fig:cm_umap} for the cases of 30\% flip noise. 
The clusters representing the classes are very well separated, and we can see that the majority of the noisy label samples have been corrected and assigned to their correct class. Similar plots for other noise types as well as critical distance plots can be found in the supplementary material~\cite{castellaniSuppl2021}.

Finally, we run \acrshort{method} on a different real-world test-set which includes a  sensor failure, and the corresponding noisy label, as shown in Fig.\ \ref{fig:CHP_data}.
The method was able to correctly re-label the period of the sensor failure.

\subsection{Ablation studies}

\subsubsection{Hyper-parameter sensitivity}
\begin{figure}[tb]
        \centering
        \includegraphics[width=0.32\textwidth]{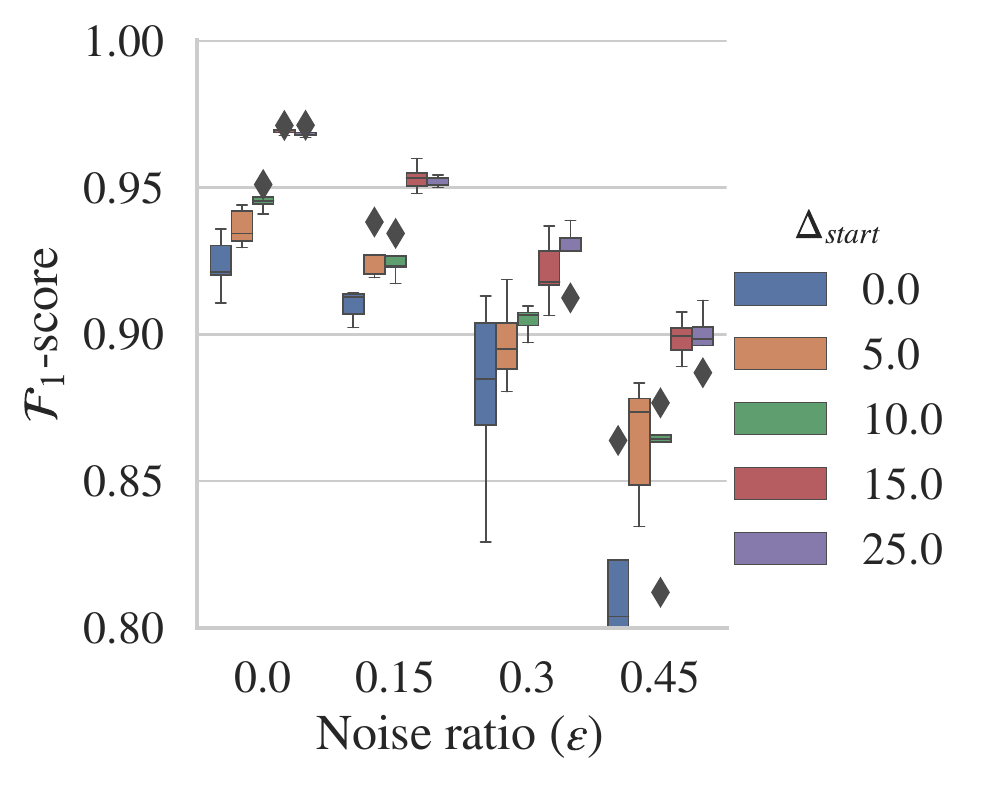}
        \includegraphics[width=0.32\textwidth]{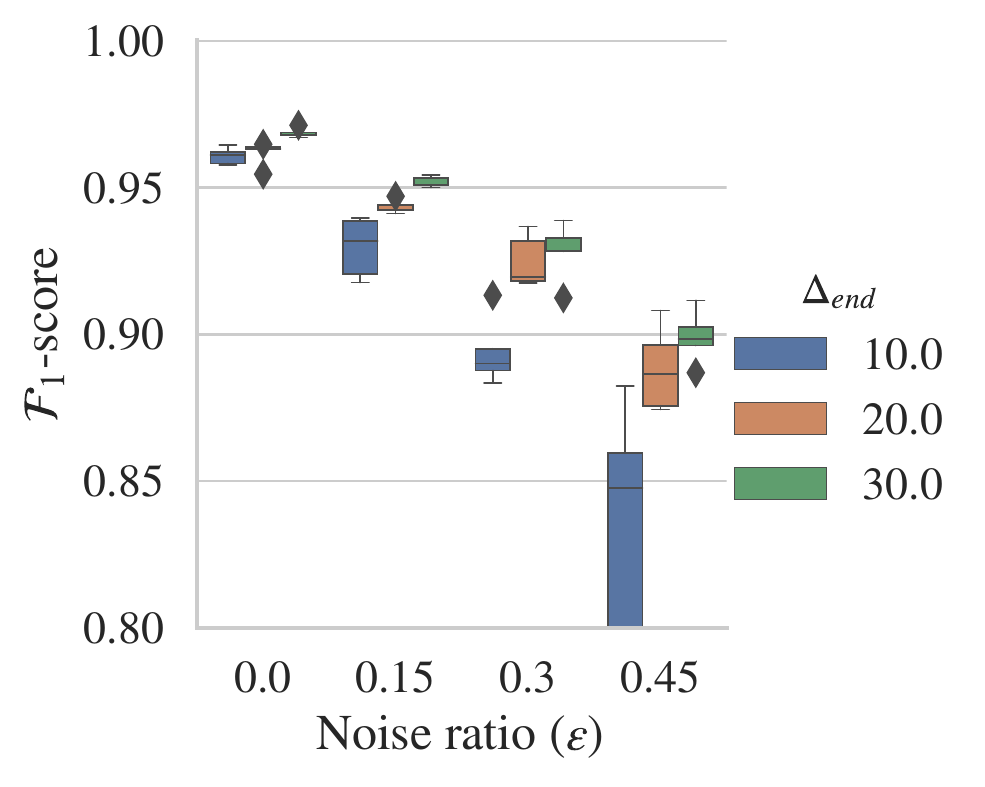}
        \includegraphics[width=0.32\textwidth]{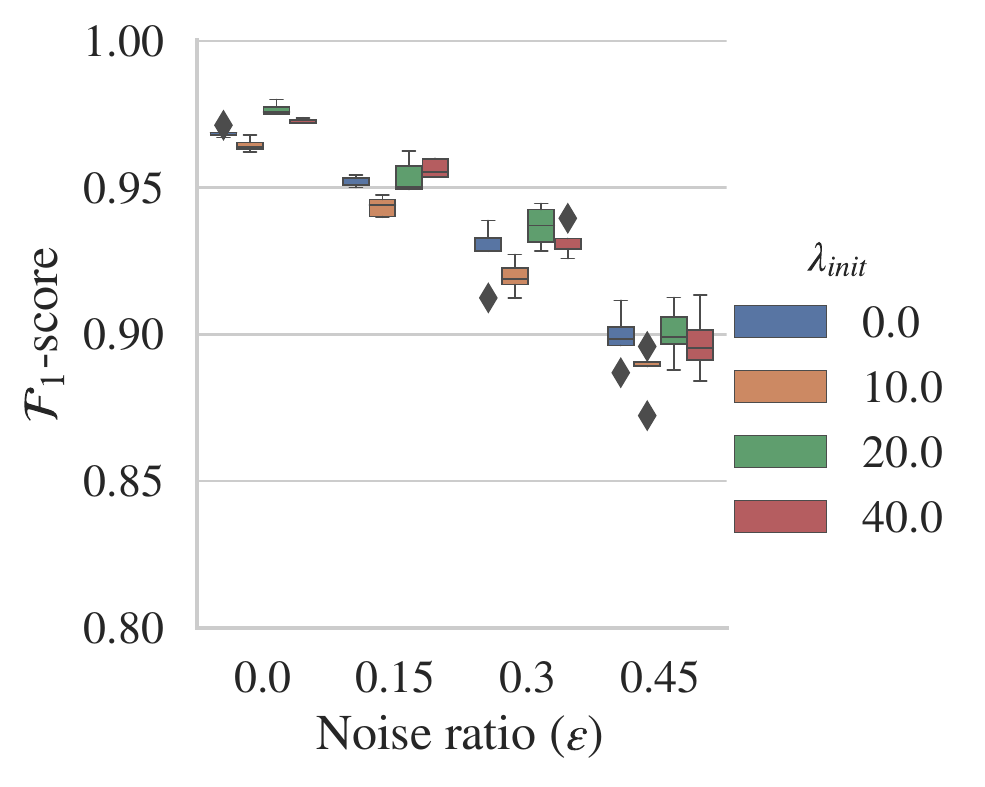}
      \caption{
      \acrshort{method} sensitivity to hyper-parameters $\Delta_{start}$ (left), $\Delta_{end}$ (middle), $\lambda_{init}$ (right) for the  CHP data and symmetric noise.}
        \label{fig:ablaton_hyper_symm}
\end{figure}

We investigate the effect of the hyper-parameters of \acrshort{method} on both, benchmarks and CHP datasets. 
The observed trends were similar in all datasets, and therefore we only report the result for the CHP dataset.
The effect of the three hyper-parameters related to the training dynamics ($\lambda_{init}$, $\Delta_{start}$, and $\Delta_{end}$) are reported in Fig.~\ref{fig:ablaton_hyper_symm} for the cases of unstructured symmetric label noise (results for the other noise types can be found in the supplementary material~\cite{castellaniSuppl2021}).
For the variation of $\Delta_{start}$  and $\Delta_{end}$ there is a clear pattern for every noise level as the performance increases with the values of the hyper-parameters, and best performance is achieved by  $\Delta_{start}=25$  and $\Delta_{end}=30$.
This shows that both the \textit{warm-up} and \textit{re-labeling} periods should be rather long and last about 55\% of the training time, before fully self-supervised training.
The effect of  $\lambda_{init}$ is not as clear, but it seems that either a random 
initialization of the cluster centers ($\lambda_{init}$ close to zero)
or an extended period of unsupervised training of the autoencoder ($\lambda_{init}$ between 20 and 40) is beneficial.

\subsubsection{Loss function components}
We study what effect each of the major loss function components of  Eq.~\ref{eq:L} has on the performance of \acrshort{method} and report the results in Table~\ref{tab:abation}.
It can be observed, that not including the constrained clustering, i.e.\ using only $\mathcal{L}_c$ or $\mathcal{L}_c+\mathcal{L}_{ae}$, gives rather poor performance for all noise types and levels. This is explicitly observed as the performance decreases again during training in the self-supervision phase without $\mathcal{L}_{cc}$  (not shown). 
This seems understandable as the label correction method repeatedly bootstraps itself by using only the labels provided by the classifier, without any anchor to preserve the information from the training labels.  
This emphasizes the necessity of constraining the data in the embedding space during self-supervision.

\begin{table*}[tb!]
    \scriptsize
    \centering
    \caption{Ablation studies on loss function components of \acrshort{method}.}
    \label{tab:abation}
    \begin{tabular*}{\textwidth}{l @{\extracolsep{\fill}} c c c c c}
        \toprule
        \textbf{Noise} & \textbf{\%} & $\bm{\mathcal{L}_c}$ & $\bm{\mathcal{L}_c + \mathcal{L}_{ae}}$ & $\bm{\mathcal{L}_c + \mathcal{L}_{cc}}$ & $\bm{\mathcal{L}_c + \mathcal{L}_{ae} + \mathcal{L}_{cc}}$\\
        \midrule
        \textbf{\textit{-}} & 0 & 0.472\rpm0.060 & 0.504\rpm0.012 & 0.974\rpm0.003 & \underline{0.980\rpm0.003} \\
        \midrule
        \multirow{3}{*}{\textbf{\textit{Symmetric}}} & 15 & 0.388\rpm0.027 & 0.429\rpm0.023 & 0.943\rpm0.004 & \underline{0.957\rpm0.007} \\
         & 30 & 0.355\rpm0.038 & 0.366\rpm0.041 & 0.919\rpm0.006 & \underline{0.930\rpm0.008} \\
         & 45 & 0.290\rpm0.014 & 0.318\rpm0.010 & 0.892\rpm0.009 & \underline{0.902\rpm0.008}\\
         \midrule
         \multirow{3}{*}{\textbf{\textit{Asymmetric}}} & 10 & 0.400\rpm0.026 & 0.407\rpm0.012 & 0.949\rpm0.003 & \underline{0.957\rpm0.003} \\
         & 20 & 0.348\rpm0.003 & 0.358\rpm0.003 & 0.930\rpm0.009 & \underline{0.941\rpm0.009}\\
         & 30 & 0.342\rpm0.002 & 0.349\rpm0.004 & 0.901\rpm0.007 & \underline{0.922\rpm0.010}\\
         \midrule
         \multirow{3}{*}{\textbf{\textit{Flip}}} & 10 & 0.460\rpm0.030 & 0.468\rpm0.030 & 0.961\rpm0.009 & \underline{0.973\rpm0.006} \\
         & 20 & 0.415\rpm0.028 & 0.424\rpm0.037 & 0.951\rpm0.008 & \underline{0.962\rpm0.005}\\
         & 30 & 0.405\rpm0.030 & 0.411\rpm0.040 & 0.943\rpm0.007 & \underline{0.957\rpm0.003}\\
         \bottomrule
    \end{tabular*}
\end{table*}

\subsubsection{Input variables}
\begin{figure}[tb!]
    \centering
    \includegraphics[width=0.9\linewidth]{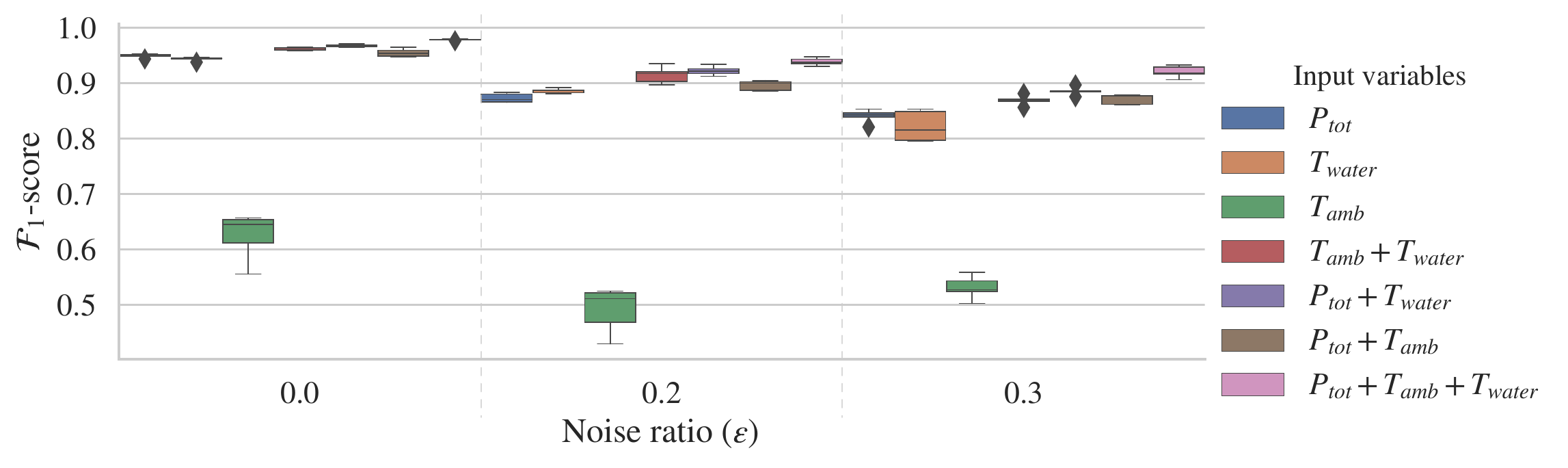}
    \caption{\acrshort{method} sensitivity to the  input variables for CHP data and asymmetric noise.}
    \label{fig:CHP_input}
\end{figure}

We investigate the influence of the selection of the input variables on the estimation of the CHP power level. The results for all possible combination of the input signals are reported in Fig~\ref{fig:CHP_input}. 
Unsurprisingly, using only the ambient temperature gives by far the worst results, while utilizing all available inputs results in the highest scores. Without the inclusion of the $P_{tot}$, we  still achieve a $\F_1$-score above 0.96 with only using the $T_{water}$ as input signal.

\section{Conclusion and future work}
In this work, we presented the problem of estimating the electrical power output of a Combined Heat and Power (CHP) machine by analyzing the total electrical power consumption of a medium size company facility. We presented an approach to estimate the CHP power output by analyzing the total load, the ambient temperature and the water temperature of the CHP, all of which are known to be control variables of the CHP. 
The training dataset for the deep-learning based approach was automatically derived from sensor measurements of the CHP power output, and sensor failures create noisy samples in the generated class labels. 
The proposed \acrfull{method} incorporates an autoencoder,  a classifier and a constraint clustering which all share and operate on a common low-dimensional embedding representation. 
During the network training, the loss function and the label correction mechanism are adjusted in a way that a robust relabeling of noisy training labels is possible.  
We compare  \acrshort{method} to five SotA label noise correction approaches on ten  time-series classification benchmarks and observe mostly comparable or better performance for various noise levels and types. 
We also observe superior performance on the CHP use-case for a wide range of noise levels and all studied noise types.
We thoroughly analyzed the dependence of the proposed methods on the (hyper-)parameters and architecture choices. 

The proposed approach is straight-forward to realize without any (hyper-)parameter tuning, as there are clear insights on how to set the parameters and the method is not sensitive to details. It also has the strong benefit that the amount of label noise need not be known or guessed. 
We used CNNs as building blocks of the proposed algorithm, but since \acrshort{method} is model agnostic, it is possible to utilize other structures, such as recurrent neural networks~\cite{Karim2018LSTMFC} or transformers~\cite{Wu2020DeepTM}, which would also utilize the time-structure of the problem. The application of such dynamic models is left for future work.

Estimating the CHP output as shown in this work will be used in the future in energy optimization scenarios to arrive at more reliable and robust EV charging schedules. 
But, due to the robustness of the proposed method and the ability to exchange the neural networks with arbitrary other machine learning modules, we see a high potential for this architecture to be used for label noise correction in other application domains.
We also see a high potential for an application in anomaly detection scenarios where sensor failures need to be detected. 
A thorough evaluation in these application areas is left for future work.

\bibliographystyle{splncs04}
\bibliography{bibliography.bib}

\end{document}


\beginsupplement

\title{Supplementary Material for: Estimating the Electrical Power Output of Industrial Devices with End-to-End Time-Series Classification in the Presence of Label Noise }
\titlerunning{Supplementary Material}

\newcommand{\orcidauthorA}{\orcidID{0000-0003-0476-5978}} 
\newcommand{\orcidauthorB}{\orcidID{0000-0001-7130-5483}} 
\newcommand{\orcidauthorC}{\orcidID{0000-0002-0935-5591}} 
\author{Andrea~Castellani \Letter \orcidauthorA \inst{1} \and
Sebastian~Schmitt \orcidauthorB \inst{2} \and
Barbara~Hammer \orcidauthorC \inst{1}} 
\authorrunning{A. Castellani and S. Schmitt and B. Hammer}
%
\institute{Bielefeld University, \\
\email{\{acastellani,bhammer\}@techfak.uni-bielefeld.de}\\ 
\and
Honda Research Institute Europe, \\
\email{sebastian.schmitt@honda-ri.de}
}

\maketitle              

\section{Implementation details}\label{app:network}
All the experiment have been conducted on a Linux Ubuntu 18.04.5 server with $2 \times $ Intel Xeon 4110 (16 cores and 32 threads), 188 GB of DDR4 RAM and $4 \times $ NVIDIA RTX 2080 Ti. The software used is Python 3.8 and the deep learning library used is PyTorch 1.4 \cite{NEURIPS2019_9015}.
\new{Our source code is publicy available at \url{https://github.com/Castel44/SREA}.}
However, we are not allowed to share the dataset used for the estimation of the CHP electrical power output because of the company privacy policy.

The network used as encoder and decoder in the experiments is a Convolutional Neural Network (CNN).
Its basic block is a convolutional layer followed by a batch normalization layer \cite{ioffe2015batch}, a non-linear activation function and a dropout layer.
The basic \textit{convolution block} (ConvBlock) is defined as:
\begin{align}
    y = &\; W \ast \bm{x} + \bm{b}  \label{eq:conv}\\
    s = &\; \text{BatchNorm}(y)  \nonumber \\
    h = &\; \text{ReLU}(s)  \nonumber \\
    o = &\; \text{Dropout}(h) \nonumber 
\end{align}

In the decoder, the convolution operation in Eq.~\ref{eq:conv} is replaced by a TransposedConvolution (equivalent to up-sampling followed by a convolution). 
Thus, the ConvBlock is called TConvBlock.

The classifier uses what so called DenseBlock, where the convolution operation in Eq.~\ref{eq:conv} is replaced by the matrix multiplication: 
\begin{equation}
    y = W \cdot \bm{x} + \bm{b}
\end{equation}
A detailed description of the network used is reported in Table~\ref{tab:network}.

In all the experiments, we train for 100 epochs in total with the Adam optimizer \cite{kingma2014adam} with weight decay of $10^{-4}$, momentum parameters set to $\beta_1=0.9$ and $\beta_2=0.999$, and \textit{eps} of $10^{-6}$ , the initial learning rate is set to 0.01. We halve the learning rate every 20 epochs. 
The dropout probability is 0.2 and the non-linear activation used is ReLU.
We use a batch size of: $\min(\frac{1}{10}\times \text{dataset\_size}, 128)$.

We stress the fact that experiments across all the datasets, noise type and ratio share the same network and hyper-parameters configuration and lead to performance on the same level and higher over the state-of-the-art (SotA).
Indeed, we are likely reporting sub-optimal results that could be improved with a noise-free validation dataset, which availability is not assumed in this work.

\new{In order to evaluate multiple classifiers, we perform the Friedman non-parametric test at 0.05 level, as described in  \cite{demvsar2006statistical}, followed by Nemenyi post-hoc test.
To visualize this comparison, we use a critical difference diagram \cite{demvsar2006statistical}, where a thick horizontal line shows the algorithm that are not-signigicantly different in terms of F-score.}

\begin{table*}[bt]
    \centering
    \caption{Neural Network structure used in this work.}
    \begin{tabular*}{\textwidth}{l @{\extracolsep{\fill}}   l l}
        \toprule
        \textit{\textbf{Name}} & \textit{\textbf{Description}} & \textit{\textbf{Dimension}} \\
        \midrule
        \multicolumn{3}{c}{\textbf{Encoder}}\\
        \midrule
        ConvBlock1 & 128 filters, $4\times1$, stride=2 & $(\text{input\_size}\times\text{seq\_len})\to(128\times\floor{\frac{\text{seq\_len}}{2}})$\\
        ConvBlock2 & 128 filters, $4\times1$, stride=2 & $(128\times\floor{\frac{\text{seq\_len}}{2}})\to(128\times\floor{\frac{\text{seq\_len}}{4}})$\\
        ConvBlock3 & 256 filters, $4\times1$, stride=2 & $(128\times\floor{\frac{\text{seq\_len}}{4}})\to(256\times\floor{\frac{\text{seq\_len}}{8}})$\\
        ConvBlock4 & 256 ch, $4\times1$, stride=2 & $(256\times\floor{\frac{\text{seq\_len}}{8}})\to(256\times\floor{\frac{\text{seq\_len}}{16}})$\\
        Pool & GlobalAveragePooling1D & $(256\times\floor{\frac{\text{seq\_len}}{16}})\to(256\times1)$\\
        \textbf{Embedding} & 32 filters, $1\times1$ & $(256\times1)\to(32\times1)$\\
        \midrule
        \multicolumn{3}{c}{\textbf{Decoder}}\\
        \midrule
        Upsample & $\floor{\frac{\text{seq\_len}}{16}}$ neurons & $(32\times1)\to(32\times\floor{\frac{\text{seq\_len}}{16}})$\\
        TConvBlock1 & 256 ch, $4\times1$, stride=2 & $(32\times\floor{\frac{\text{seq\_len}}{16}})\to(256\times\floor{\frac{\text{seq\_len}}{16}})$\\
        TConvBlock2 & 256 filters, $4\times1$, stride=2 & $(256\times\floor{\frac{\text{seq\_len}}{8}})\to(128\times\floor{\frac{\text{seq\_len}}{4}})$\\
        TConvBlock3 & 128 filters, $4\times1$, stride=2 & $(128\times\floor{\frac{\text{seq\_len}}{4}})\to(128\times\floor{\frac{\text{seq\_len}}{2}})$\\
        TConvBlock4 & 128 filters, $4\times1$, stride=2 & $(128\times\floor{\frac{\text{seq\_len}}{2}})\to(\text{input\_size}\times\text{seq\_len})$\\
        \midrule
        \multicolumn{3}{c}{\textbf{Classifier}}\\
        \midrule
        DenseBlock & 128 neurons & $(32\times1)\to(128\times1)$\\
        Output & $\#$class neurons & $(128\times1)\to(\text{\#class}\times1)$\\
         \bottomrule
    \end{tabular*}
    \label{tab:network}
\end{table*}

\clearpage
\section{Definition of noise}\label{app:T}
The definition of the labels \textit{transition matrix} T, being $\epsilon \in [0, 1]$ the noise ratio and $k$ the number of classes, is as follows:

\begin{align*}
    \text{Symmetric noise: } T= & \,
    \begin{bmatrix}
        1 - \epsilon & \frac{\epsilon}{k -1} & \cdots & \frac{\epsilon}{k -1} & \frac{\epsilon}{k -1} \\
        \frac{\epsilon}{k -1} & 1-\epsilon & \frac{\epsilon}{k -1} & \cdots & \frac{\epsilon}{k -1}\\
        \vdots & & \ddots & & \vdots\\
        \frac{\epsilon}{k -1} & \cdots & \frac{\epsilon}{k -1} & 1-\epsilon & \frac{\epsilon}{k -1}\\
        \frac{\epsilon}{k -1} & \frac{\epsilon}{k -1} & \cdots & \frac{\epsilon}{k -1} & 1 - \epsilon
    \end{bmatrix}\\
    \text{Asymmetric noise: } T= &\,
    \begin{bmatrix}
        1-\epsilon & \epsilon & 0 & \cdots & 0 \\
        0 & 1-\epsilon & \epsilon & & 0\\
        \vdots & & \ddots & \ddots & \vdots\\
        0 & & & 1-\epsilon & \epsilon\\
        \epsilon & 0 & \cdots & 0 & 1-\epsilon
    \end{bmatrix}\\
    \text{Pair noise: } T = & \,
    \begin{bmatrix}
        1 & 0 & \cdots & & 0 \\
        \epsilon & 1-\epsilon & 0 & \cdots & 0 \\
        \vdots & & \ddots & & \vdots \\
        \epsilon & 0 & \cdots & 0 & 1-\epsilon
    \end{bmatrix}
\end{align*}

\section{Benchmarks Datasets results}
\begin{figure}[th]
    \centering
    \includegraphics[width=0.8\textwidth]{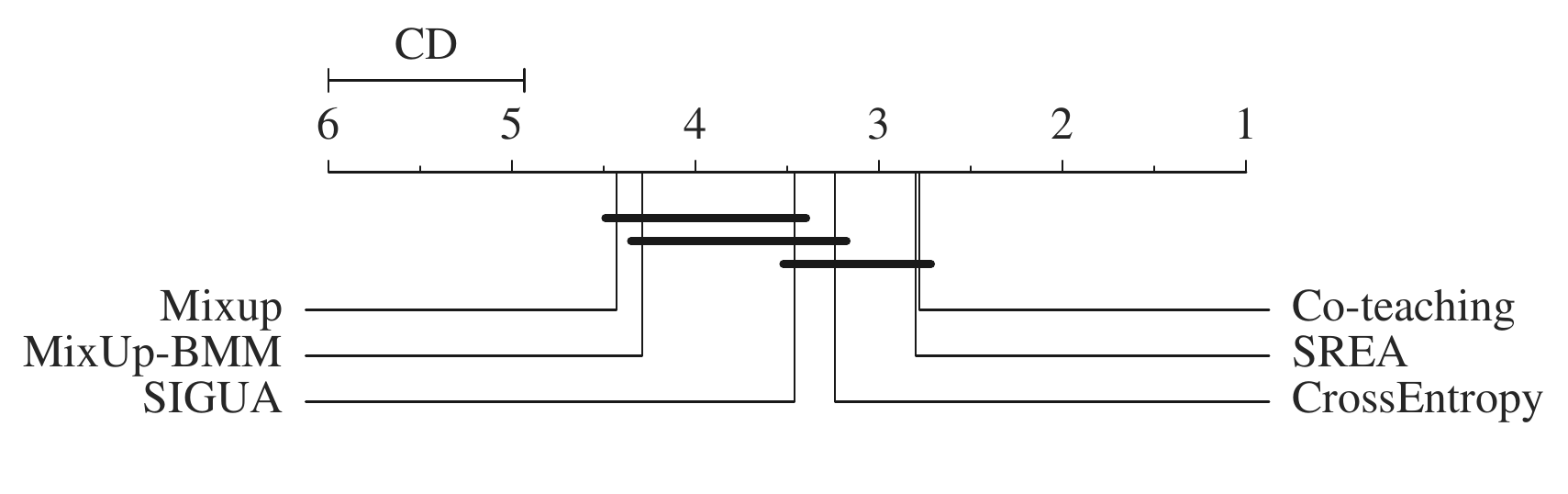}
    \caption{Critical difference diagram showing pairwise statistical difference comparison of \acrshort{method} and SoTA on UCR benchmark datasets corrupted with symmetric label noise.}
    \label{fig:UCR_CD_symm}
\end{figure}

\begin{figure}[th]
    \centering
    \includegraphics[width=0.8\textwidth]{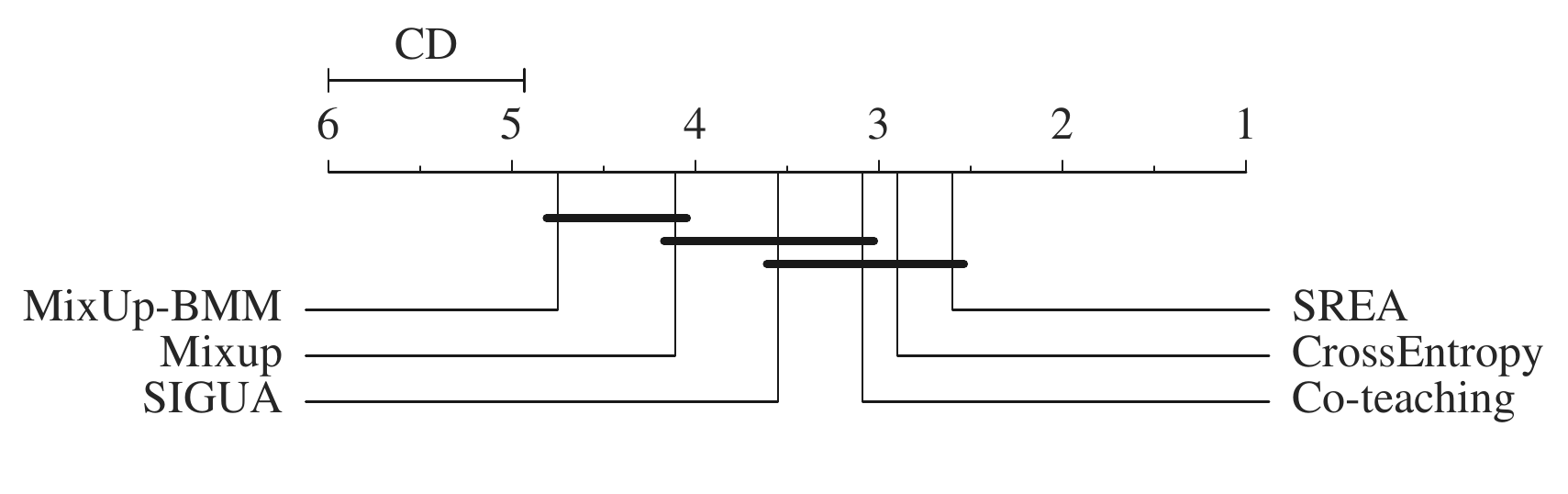}
    \caption{Critical difference diagram showing pairwise statistical difference comparison of \acrshort{method} and SoTA on UCR benchmark datasets corrupted with asymmetric label noise.}
    \label{fig:UCR_CD_asymm}
\end{figure}

\begin{figure}[th]
    \centering
    \includegraphics[width=0.8\textwidth]{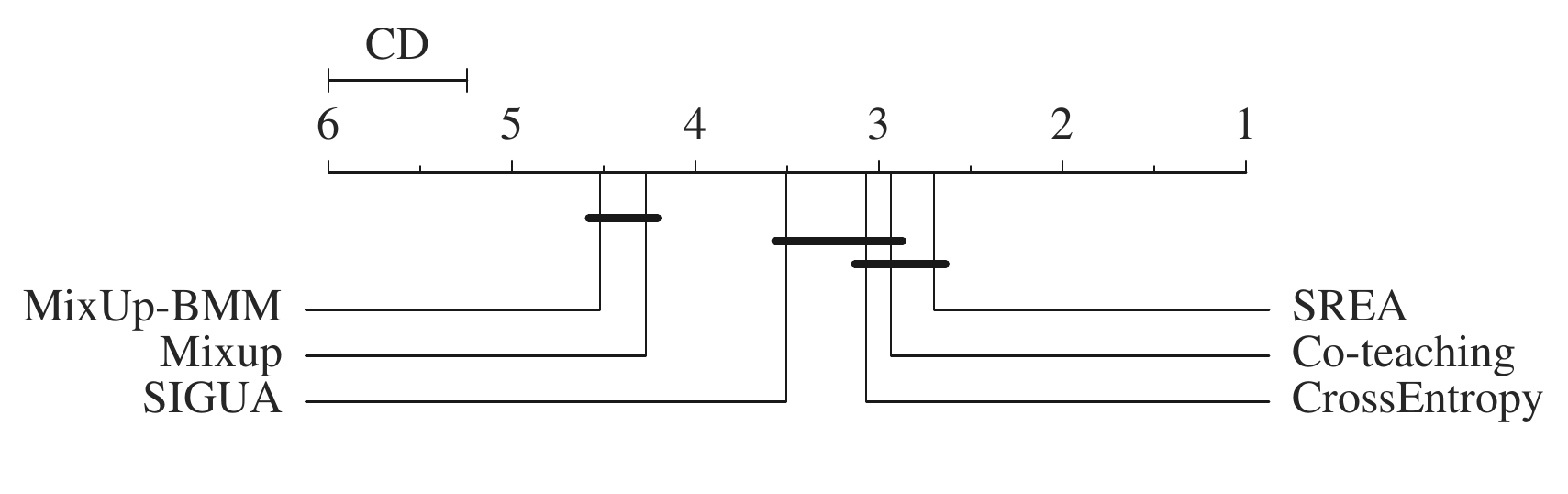}
    \caption{Critical difference diagram showing pairwise statistical difference comparison of \acrshort{method} and SoTA on UCR benchmark datasets corrupted with both symmetric and asymmetric label noise.}
    \label{fig:UCR_CD}
\end{figure}

In Table~\ref{tab:ucr_symm} and Table~\ref{tab:ucr_asymm} we report the complete set of results with the 10 selected dataset from the UCR repository \cite{UCRArchive2018}, under presence of symmetric and asymmetric noise respectively.
\new{In the figures \ref{fig:UCR_CD_symm}, \ref{fig:UCR_CD_asymm} and \ref{fig:UCR_CD} is shown the critical difference diagram over the UCR benchmarks dataset under presence of symmetric, asymmetric and both lable noise respectively.}

\begin{table*}[h!]
    \centering
    \scriptsize
    \caption{$\F_1$ test scores on UCR datasets with symmetric noise.
    \new{In parenthesis the results of a Mann–Whitney U test with $\alpha=0.05$ of \acrshort{method} against the other approachs: SREA $\F_1$ is significantly higher ($+$), lower ($-$) or not significant ($\approx$).}}
    \label{tab:ucr_symm}
    \begin{tabular*}{\linewidth}{l @{\extracolsep{\fill}} l l l l l l l}
    \toprule
    \textbf{Dataset}&\textbf{\%}&\textbf{CE}&\textbf{MixUp}&\textbf{M-BMM}&\textbf{SIGUA}&\textbf{Co-teach}&\textbf{\acrshort{method}}\\
    \midrule
    \multirow{5}{0.13\linewidth}{ArrowHead}
    & 0     & 0.899 $(\approx)$  & 0.841 $(\approx)$ & 0.751 $(+)$ & 0.859 $(\approx)$ & 0.819 $(+)$ & 0.855 \\
    & 15  & 0.813 $(\approx)$ & 0.723 $(\approx)$ & 0.721 $(\approx)$ & 0.700 $(\approx)$ & 0.766 $(\approx)$ & 0.751\\
    & 30  & 0.721 $(\approx)$ & 0.679 $(\approx)$ & 0.632 $(\approx)$ & 0.626 $(\approx)$ & 0.734 $(\approx)$ & 0.651\\
    & 45 & 0.569 $(-)$ & 0.509 $(\approx)$ & 0.517 $(\approx)$ & 0.527 $(\approx)$ & 0.533 $(\approx)$ & 0.445\\
    & 60 & 0.394 $(-)$ & 0.435 $(-)$ & 0.386 $(\approx)$ & 0.370 $(-)$ & 0.405 $(-)$ & 0.295 \\
    \midrule
    \multirow{5}{0.13\linewidth}{CBF}
    & 0     & 1.000 $(+)$ & 0.970 $(+)$ & 0.886 $(+)$ & 1.000 $(+)$ & 0.997 $(+)$ & 1.000\\
    & 15  & 0.943 $(+)$ & 0.923 $(+)$ & 0.941 $(+)$ & 0.976 $(+)$ & 0.923 $(+)$ & 1.000 \\
    & 30  & 0.780 $(+)$ & 0.799 $(+)$ & 0.932 $(+)$ & 0.923 $(+)$ & 0.833 $(+)$ & 0.998 \\
    & 45  & 0.570 $(+)$ & 0.635 $(+)$ & 0.846 $(+)$ & 0.730 $(+)$ & 0.617 $(+)$ & 0.981 \\
    & 60   & 0.450 $(\approx)$ & 0.456 $(\approx)$ & 0.552 $(\approx)$ & 0.390 $(\approx)$ & 0.448 $(\approx)$ & 0.347 \\
    \midrule
    \multirow{5}{0.13\linewidth}{Epilepsy}
    & 0 & 0.974 $(\approx)$ & 0.955 $(+)$ & 0.926 $(+)$ & 0.978 $(\approx)$ & 0.971 $(+)$ & 0.973 \\
    & 15 & 0.890 $(\approx)$ & 0.913 $(\approx)$ & 0.899 $(\approx)$ & 0.884 $(\approx)$ & 0.861 $(\approx)$ & 0.861\\
    & 30 & 0.784 $(-)$ & 0.823 $(-)$ & 0.805 $(-)$ &0.741 $(\approx)$ &0.744 $(\approx)$ & 0.708 \\
    & 45 & 0.604 $(-)$ & 0.630 $(-)$ & 0.650 $(-)$ & 0.600 $(-)$ & 0.661 $(-)$ & 0.446 \\
    & 60 & 0.464 $(-)$ & 0.441 $(-)$ & 0.446 $(-)$ & 0.459 $(-)$ & 0.399 $(\approx)$ & 0.340\\
    \midrule
    \multirow{5}{0.13\linewidth}{FaceFour}
    & 0 & 0.991 $(\approx)$ & 0.974 $(+)$ & 0.929 $(+)$ & 0.955 $(+)$ & 0.908 $(+)$ & 1.000 \\
    & 15 & 0.893 $(\approx)$ & 0.836 $(+)$ & 0.834 $(+)$ & 0.845 $(\approx)$ & 0.866 $(\approx)$ & 0.931\\
    & 30 & 0.667 $(\approx)$ & 0.658 $(\approx)$ & 0.699 $(\approx)$ & 0.719 $(-)$ & 0.706 $(\approx)$ & 0.663\\
    & 45 & 0.498 $(\approx)$ & 0.493 $(\approx)$ & 0.470 $(\approx)$ & 0.455 $(\approx)$ & 0.504 $(\approx)$ & 0.556\\
    & 60 & 0.364 $(\approx)$ & 0.314 $(\approx)$ & 0.390 $(\approx)$ & 0.274 $(\approx)$ & 0.411 $(\approx)$ & 0.342\\
    \midrule
    \multirow{5}{0.13\linewidth}{Melbourne}
    & 0 & 0.923 $(\approx)$ & 0.879 $(+)$ & 0.773 $(+)$ & 0.918 $(\approx)$ & 0.913 $(\approx)$ & 0.911\\
    & 15 & 0.869 $(+)$ & 0.870 $(+)$ & 0.856 $(+)$ & 0.883 $(\approx)$ & 0.886 $(\approx)$ & 0.883\\
    & 30 & 0.826 $(+)$ & 0.858 $(\approx)$ & 0.870 $(\approx)$ & 0.855 $(\approx)$ & 0.876 $(-)$ & 0.862\\
    & 45 & 0.767 $(+)$ & 0.817 $(+)$ & 0.818 $(\approx)$ & 0.832 $(\approx)$ & 0.848 $(\approx)$ & 0.841\\
    & 60 & 0.662 $(\approx)$ & 0.716 $(+)$ & 0.734 $(+)$ & 0.778 $(+)$ & 0.805 $(+)$ & 0.812\\
    \midrule
    \multirow{5}{0.13\linewidth}{NATOPS}
    & 0 & 0.858 $(\approx)$ & 0.801 $(\approx)$ & 0.711 $(+)$ & 0.848 $(\approx)$ & 0.835 $(\approx)$ & 0.866\\
    & 15 & 0.779 $(\approx)$ & 0.718 $(+)$ & 0.702 $(+)$ & 0.754 $(\approx)$ & 0.761 $(\approx)$ & 0.796\\
    & 30 & 0.587 $(\approx)$ & 0.580 $(+)$ & 0.602 $(+)$ & 0.593 $(+)$ & 0.673 $(+)$ & 0.670\\
    & 45 & 0.436 $(\approx)$ & 0.466 $(\approx)$ & 0.544 $(\approx)$ & 0.452 $(\approx)$ & 0.516 $(\approx)$ & 0.483\\
    & 60 & 0.320 $(\approx)$ & 0.303 $(\approx)$ & 0.339 $(\approx)$ & 0.328 $(\approx)$ & 0.410 $(-)$ & 0.335\\
    \midrule
    \multirow{5}{0.13\linewidth}{OSULeaf}
    & 0 & 0.852 $(\approx)$ & 0.590 $(\approx)$ & 0.478 $(+)$ & 0.796 $(\approx)$ & 0.828 $(\approx)$ & 0.706\\
    & 15 & 0.781 $(-)$ & 0.558 $(\approx)$ & 0.502 $(\approx)$ & 0.694 $(\approx)$ & 0.741 $(\approx)$ & 0.651\\
    & 30 & 0.692 $(\approx)$ & 0.400 $(+)$ & 0.553 $(\approx)$ & 0.611 $(\approx)$ & 0.673 $(\approx)$ & 0.554\\
    & 45 & 0.534 $(-)$ & 0.384 $(\approx)$ & 0.400 $(\approx)$ & 0.474 $(-)$ & 0.539 $(-)$ & 0.355\\
    & 60 & 0.427 $(-)$ & 0.305 $(-)$ & 0.328 $(-)$ & 0.376 $(-)$ & 0.420 $(-)$ & 0.252\\
    \midrule
    \multirow{5}{0.13\linewidth}{Plane}
    & 0 & 0.995 $(\approx)$ & 0.962 $(+)$ & 0.577 $(+)$ & 0.981 $(+)$ & 0.990 $(\approx)$ & 0.998\\
    & 15 & 0.930 $(+)$ & 0.953 $(+)$ & 0.873 $(+)$ & 0.971 $(\approx)$ & 0.981 $(\approx)$ & 0.983\\
    & 30 & 0.887 $(+)$ & 0.902 $(+)$ & 0.943 $(\approx)$ & 0.862 $(+)$ & 0.941 $(\approx)$ & 0.944\\
    & 45 & 0.671 $(+)$ & 0.783 $(+)$ & 0.761 $(+)$ & 0.797 $(+)$ & 0.808 $(+)$ & 0.911\\
    & 60 & 0.575 $(\approx)$ & 0.593 $(\approx)$ & 0.561 $(\approx)$ & 0.529 $(\approx)$ & 0.659 $(\approx)$ & 0.629\\
    \midrule
    \multirow{5}{0.13\linewidth}{Symbols}
    & 0 & 0.994 $(-)$ & 0.931 $(+)$ & 0.642 $(+)$ & 0.979 $(\approx)$ & 0.979 $(\approx)$ & 0.987\\ 
    & 15 & 0.982 $(\approx)$ & 0.871 $(+)$ & 0.689 $(+)$ & 0.968 $(\approx)$ & 0.979 $(\approx)$ & 0.983\\
    & 30 & 0.974 $(\approx)$ & 0.880 $(+)$ & 0.911 $(+)$ & 0.940 $(+)$ & 0.982 $(\approx)$ & 0.983\\
    & 45 & 0.945 $(\approx)$ & 0.815 $(\approx)$ & 0.839 $(\approx)$ & 0.916 $(\approx)$ & 0.928 $(\approx)$ & 0.891\\
    & 60 & 0.862 $(-)$ & 0.738 $(-)$ & 0.599 $(\approx)$ & 0.790 $(-)$ & 0.945 $(-)$ & 0.326\\
    \midrule
    \multirow{5}{0.13\linewidth}{Trace}
    & 0 & 1.000 $(\approx)$ & 0.995 $(\approx)$ & 0.634 $(+)$ & 1.000 $(\approx)$ & 0.933 $(+)$ & 1.000\\
    & 15 & 0.903 $(+)$ & 0.870 $(+)$ & 0.866 $(+)$ & 0.980 $(+)$ & 0.965 $(+)$ & 0.995\\
    & 30 & 0.702 $(+)$ & 0.773 $(+)$ & 0.629 $(+)$ & 0.892 $(+)$ & 0.887 $(+)$ & 0.952\\
    & 45 & 0.612 $(\approx)$ & 0.518 $(+)$ & 0.643 $(\approx)$ & 0.646 $(\approx)$ & 0.781 $(\approx)$ & 0.667\\
    & 60 & 0.402 $(\approx)$ & 0.347 $(+)$ & 0.435 $(\approx)$ & 0.378 $(\approx)$ & 0.552 $(\approx)$ & 0.559\\
    \bottomrule
    \toprule
    \multirow{6}{0.13\linewidth}{\textit{\textbf{Summary Results}}}
    & \textbf{\%} & \multicolumn{2}{c}{\textit{\textbf{\#Better $(+)$}}} & \multicolumn{2}{c}{\textit{\textbf{\#Equal $(\approx)$}}} & \multicolumn{2}{c}{\textit{\textbf{\#Worse $(-)$}}} \\
    & 0 &\multicolumn{2}{c}{\textit{\textbf{26}}} & \multicolumn{2}{c}{\textit{\textbf{23}}} & \multicolumn{2}{c}{\textit{\textbf{1}}} \\
    & 15 &\multicolumn{2}{c}{\textit{\textbf{22}}} & \multicolumn{2}{c}{\textit{\textbf{27}}} & \multicolumn{2}{c}{\textit{\textbf{1}}} \\
    & 30 &\multicolumn{2}{c}{\textit{\textbf{22}}} & \multicolumn{2}{c}{\textit{\textbf{23}}} & \multicolumn{2}{c}{\textit{\textbf{5}}} \\
    & 45 &\multicolumn{2}{c}{\textit{\textbf{13}}} & \multicolumn{2}{c}{\textit{\textbf{28}}} & \multicolumn{2}{c}{\textit{\textbf{9}}} \\
    & 60 &\multicolumn{2}{c}{\textit{\textbf{5}}} & \multicolumn{2}{c}{\textit{\textbf{27}}} & \multicolumn{2}{c}{\textit{\textbf{18}}} \\
    \bottomrule
    \end{tabular*}

\end{table*}

\begin{table*}[th!]
    \centering
    \scriptsize
    \caption{$\mathcal{F}_1$ test scores on UCR datasets with asymmetric noise.
    \new{In parenthesis the results of a Mann–Whitney U test with $\alpha=0.05$ of \acrshort{method} against the other approachs: SREA $\F_1$ is significantly higher ($+$), lower ($-$) or not significant ($\approx$).}}
    \label{tab:ucr_asymm}
    \begin{tabular*}{\linewidth}{l @{\extracolsep{\fill}} l l l l l l l}
    \toprule
    \textbf{Dataset}&\textbf{\%}&\textbf{CE}&\textbf{MixUp}&\textbf{M-BMM}&\textbf{SIGUA}&\textbf{Co-teach}&\textbf{\acrshort{method}}\\
    \midrule
    \multirow{5}{0.13\linewidth}{ArrowHead}
    & 0   & 0.895 $(\approx)$  & 0.850 $(\approx)$ & 0.761 $(+)$ & 0.867 $(\approx)$ & 0.835 $(+)$ & 0.864 \\
    & 10  & 0.850 $(\approx)$ & 0.779 $(\approx)$ & 0.712 $(\approx)$ & 0.743 $(+)$ & 0.825 $(\approx)$ & 0.821\\
    & 20  & 0.790 $(\approx)$ & 0.729 $(\approx)$ & 0.685 $(\approx)$ & 0.724 $(\approx)$ & 0.748 $(\approx)$ & 0.753\\
    & 30  & 0.741 $(\approx)$ & 0.685 $(\approx)$ & 0.616 $(\approx)$ & 0.609 $(\approx)$ & 0.699 $(\approx)$ & 0.563\\
    & 40 & 0.644 $(\approx)$ & 0.592 $(\approx)$ & 0.573 $(\approx)$ & 0.592 $(\approx)$ & 0.512 $(\approx)$ & 0.498 \\
    \midrule
    \multirow{5}{0.13\linewidth}{CBF}
    & 0     & 1.000 $(+)$ & 0.965 $(+)$ & 0.915 $(+)$ & 1.000 $(+)$ & 1.000 $(+)$ & 1.000\\
    & 10  & 0.973 $(+)$ & 0.956 $(+)$ & 0.920 $(+)$ & 0.989 $(+)$ & 0.963 $(+)$ & 1.000 \\
    & 20  & 0.905 $(+)$ & 0.897 $(+)$ & 0.949 $(+)$ & 0.980 $(+)$ & 0.900 $(+)$ & 1.000 \\
    & 30  & 0.779 $(+)$ & 0.771 $(+)$ & 0.954 $(+)$ & 0.880 $(+)$ & 0.857 $(+)$ & 0.999 \\
    & 40   & 0.663 $(+)$ & 0.644 $(+)$ & 0.816 $(+)$ & 0.681 $(+)$ & 0.665 $(+)$ & 0.989 \\
    \midrule
    \multirow{5}{0.13\linewidth}{Epilepsy}
    & 0 & 0.978 $(\approx)$ & 0.951 $(+)$ & 0.905 $(+)$ & 0.948 $(+)$ & 0.956 $(+)$ & 0.984 \\
    & 10 & 0.919 $(\approx)$ & 0.930 $(\approx)$ & 0.847 $(\approx)$ & 0.905 $(\approx)$ & 0.919 $(\approx)$ & 0.888\\
    & 20 & 0.861 $(\approx)$ & 0.894 $(-)$ & 0.891 $(-)$ &0.826 $(\approx)$ &0.863 $(\approx)$ & 0.825 \\
    & 30 & 0.781 $(-)$ & 0.783 $(-)$ & 0.800 $(-)$ & 0.783 $(\approx)$ & 0.727 $(\approx)$ & 0.712 \\
    & 40 & 0.648 $(-)$ & 0.688 $(-)$ & 0.647 $(-)$ & 0.642 $(\approx)$ & 0.671 $(-)$ & 0.571\\
    \midrule
    \multirow{5}{0.13\linewidth}{FaceFour}
    & 0 & 0.974 $(\approx)$ & 0.956 $(+)$ & 0.965 $(+)$ & 0.945 $(+)$ & 0.887 $(+)$ & 0.974 \\
    & 10 & 0.900 $(\approx)$ & 0.844 $(+)$ & 0.879 $(\approx)$ & 0.864 $(+)$ & 0.855 $(+)$ & 0.939\\
    & 20 & 0.754 $(+)$ & 0.720 $(+)$ & 0.748 $(+)$ & 0.721 $(+)$ & 0.696 $(+)$ & 0.912\\
    & 30 & 0.630 $(+)$ & 0.639 $(+)$ & 0.624 $(+)$ & 0.657 $(+)$ & 0.726 $(\approx)$ & 0.811\\
    & 40 & 0.568 $(\approx)$ & 0.516 $(\approx)$ & 0.546 $(\approx)$ & 0.450 $(\approx)$ & 0.613 $(\approx)$ & 0.686\\
    \midrule
    \multirow{5}{0.13\linewidth}{Melbourne}
    & 0 & 0.921 $(\approx)$ & 0.878 $(+)$ & 0.744 $(+)$ & 0.913 $(\approx)$ & 0.911 $(\approx)$ & 0.923\\
    & 10 & 0.898 $(+)$ & 0.877 $(+)$ & 0.860 $(+)$ & 0.899 $(+)$ & 0.897 $(+)$ & 0.911\\
    & 20 & 0.865 $(+)$ & 0.861 $(+)$ & 0.851 $(+)$ & 0.858 $(+)$ & 0.893 $(\approx)$ & 0.903\\
    & 30 & 0.790 $(+)$ & 0.839 $(+)$ & 0.826 $(+)$ & 0.829 $(+)$ & 0.864 $(+)$ & 0.889\\
    & 40 & 0.701 $(+)$ & 0.757 $(+)$ & 0.689 $(+)$ & 0.792 $(+)$ & 0.831 $(\approx)$ & 0.836\\
    \midrule
    \multirow{5}{0.13\linewidth}{NATOPS}
    & 0 & 0.843 $(\approx)$ & 0.818 $(\approx)$ & 0.724 $(+)$ & 0.842 $(\approx)$ & 0.841 $(\approx)$ & 0.851\\
    & 10 & 0.798 $(+)$ & 0.822 $(\approx)$ & 0.756 $(+)$ & 0.764 $(+)$ & 0.790 $(+)$ & 0.829\\
    & 20 & 0.703 $(\approx)$ & 0.763 $(\approx)$ & 0.762 $(\approx)$ & 0.698 $(\approx)$ & 0.733 $(\approx)$ & 0.762\\
    & 30 & 0.627 $(\approx)$ & 0.678 $(\approx)$ & 0.695 $(\approx)$ & 0.654 $(\approx)$ & 0.644 $(\approx)$ & 0.675\\
    & 40 & 0.503 $(\approx)$ & 0.586 $(\approx)$ & 0.609 $(\approx)$ & 0.443 $(+)$ & 0.562 $(\approx)$ & 0.576\\
    \midrule
    \multirow{5}{0.13\linewidth}{OSULeaf}
    & 0 & 0.862 $(\approx)$ & 0.582 $(\approx)$ & 0.579 $(+)$ & 0.806 $(\approx)$ & 0.865 $(\approx)$ & 0.751\\
    & 10 & 0.796 $(\approx)$ & 0.575 $(\approx)$ & 0.550 $(+)$ & 0.760 $(\approx)$ & 0.805 $(\approx)$ & 0.686\\
    & 20 & 0.730 $(\approx)$ & 0.473 $(\approx)$ & 0.502 $(\approx)$ & 0.680 $(\approx)$ & 0.718 $(\approx)$ & 0.634\\
    & 30 & 0.654 $(-)$ & 0.499 $(\approx)$ & 0.427 $(\approx)$ & 0.587 $(\approx)$ & 0.676 $(-)$ & 0.533\\
    & 40 & 0.520 $(-)$ & 0.354 $(-)$ & 0.327 $(-)$ & 0.450 $(-)$ & 0.467 $(-)$ & 0.355\\
    \midrule
    \multirow{5}{0.13\linewidth}{Plane}
    & 0 & 0.995 $(\approx)$ & 0.962 $(+)$ & 0.619 $(+)$ & 0.986 $(+)$ & 0.995 $(\approx)$ & 0.995\\
    & 10 & 0.981 $(\approx)$ & 0.986 $(\approx)$ & 0.648 $(+)$ & 0.986 $(\approx)$ & 0.990 $(\approx)$ & 0.976\\
    & 20 & 0.952 $(\approx)$ & 0.923 $(\approx)$ & 0.751 $(+)$ & 0.976 $(\approx)$ & 0.990 $(\approx)$ & 0.966\\
    & 30 & 0.886 $(+)$ & 0.877 $(\approx)$ & 0.686 $(+)$ & 0.899 $(\approx)$ & 0.924 $(\approx)$ & 0.941\\
    & 40 & 0.745 $(\approx)$ & 0.741 $(\approx)$ & 0.637 $(\approx)$ & 0.694 $(+)$ & 0.738 $(\approx)$ & 0.824\\
    \midrule
    \multirow{5}{0.13\linewidth}{Symbols}
    & 0 & 0.992 $(-)$ & 0.941 $(+)$ & 0.684 $(+)$ & 0.988 $(\approx)$ & 0.987 $(\approx)$ & 0.980\\ 
    & 10 & 0.989 $(-)$ & 0.902 $(+)$ & 0.676 $(+)$ & 0.981 $(\approx)$ & 0.986 $(-)$ & 0.980\\
    & 20 & 0.990 $(-)$ & 0.907 $(+)$ & 0.677 $(+)$ & 0.985 $(-)$ & 0.958 $(+)$ & 0.977\\
    & 30 & 0.984 $(-)$ & 0.707 $(+)$ & 0.696 $(+)$ & 0.857 $(+)$ & 0.968 $(\approx)$ & 0.955\\
    & 40 & 0.908 $(\approx)$ & 0.594 $(+)$ & 0.573 $(+)$ & 0.791 $(\approx)$ & 0.916 $(\approx)$ & 0.798\\
    \midrule
    \multirow{5}{0.13\linewidth}{Trace}
    & 0 & 1.000 $(\approx)$ & 1.000 $(\approx)$ & 0.704 $(+)$ & 0.984 $(\approx)$ & 0.962 $(+)$ & 1.000\\
    & 10 & 0.934 $(+)$ & 0.758 $(+)$ & 0.781 $(+)$ & 0.953 $(\approx)$ & 0.891 $(+)$ & 0.992\\
    & 20 & 0.899 $(+)$ & 0.769 $(+)$ & 0.588 $(+)$ & 0.912 $(+)$ & 0.919 $(+)$ & 0.990\\
    & 30 & 0.737 $(+)$ & 0.665 $(+)$ & 0.551 $(+)$ & 0.833 $(\approx)$ & 0.849 $(\approx)$ & 0.876\\
    & 40 & 0.629 $(\approx)$ & 0.576 $(\approx)$ & 0.549 $(+)$ & 0.467 $(+)$ & 0.679 $(\approx)$ & 0.726\\
    \bottomrule
    \toprule
    \multirow{6}{0.13\linewidth}{\textit{\textbf{Summary Results}}}
    & \textbf{\%} & \multicolumn{2}{c}{\textit{\textbf{\#Better $(+)$}}} & \multicolumn{2}{c}{\textit{\textbf{\#Equal $(\approx)$}}} & \multicolumn{2}{c}{\textit{\textbf{\#Worse $(-)$}}} \\
    & 0 &\multicolumn{2}{c}{\textit{\textbf{26}}} & \multicolumn{2}{c}{\textit{\textbf{23}}} & \multicolumn{2}{c}{\textit{\textbf{1}}} \\
    & 10 &\multicolumn{2}{c}{\textit{\textbf{26}}} & \multicolumn{2}{c}{\textit{\textbf{22}}} & \multicolumn{2}{c}{\textit{\textbf{2}}} \\
    & 20 &\multicolumn{2}{c}{\textit{\textbf{23}}} & \multicolumn{2}{c}{\textit{\textbf{23}}} & \multicolumn{2}{c}{\textit{\textbf{4}}} \\
    & 30 &\multicolumn{2}{c}{\textit{\textbf{22}}} & \multicolumn{2}{c}{\textit{\textbf{22}}} & \multicolumn{2}{c}{\textit{\textbf{6}}} \\
    & 40 &\multicolumn{2}{c}{\textit{\textbf{15}}} & \multicolumn{2}{c}{\textit{\textbf{30}}} & \multicolumn{2}{c}{\textit{\textbf{5}}} \\
    \bottomrule
    \end{tabular*}

\end{table*}

\clearpage
\section{CHP electrical power output estimation results}
\begin{figure}[tb]
    \centering
    \includegraphics[width=0.80\textwidth]{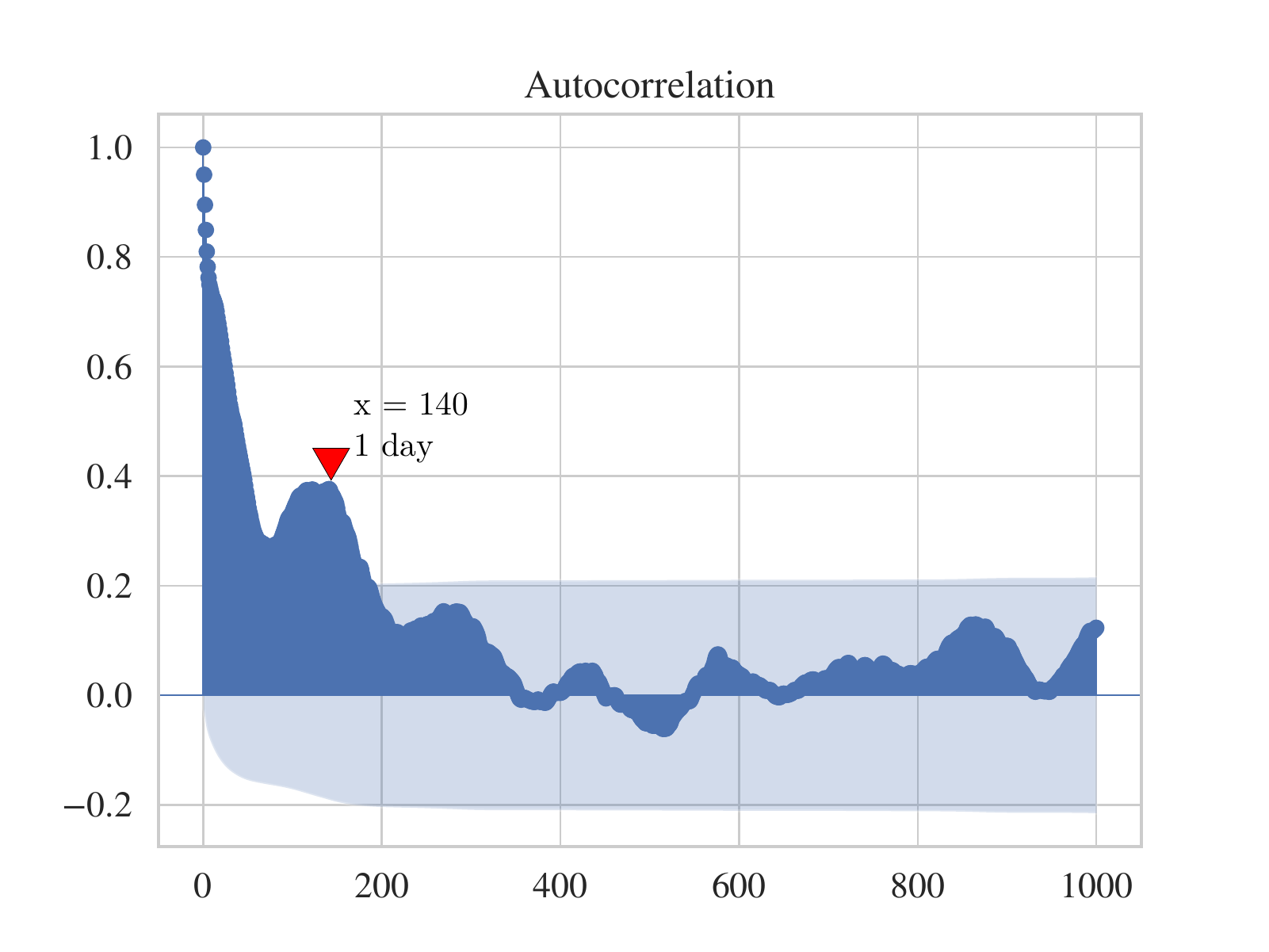}
    \caption{Autocorrelation plot for the CHP dataset. The autocorrelation function has the second-largest positive peak at 140 samples (1 day).}
    \label{fig:ACF}
\end{figure}

\new{In Fig.~\ref{fig:ACF} is shown the plot of the autocorrelation function (ACF) on the CHP dataset. It is clearly notable a positive peak at 140 samples, which means that there is a daily seasonal period.
Please note that the time-structure of the time series data is not utilized in this work. Each sample is considered completely independent from each other, even though they might be derived from two adjacent sliding windows.}

\begin{figure}[th]
    \centering
    \includegraphics[width=0.4\textwidth]{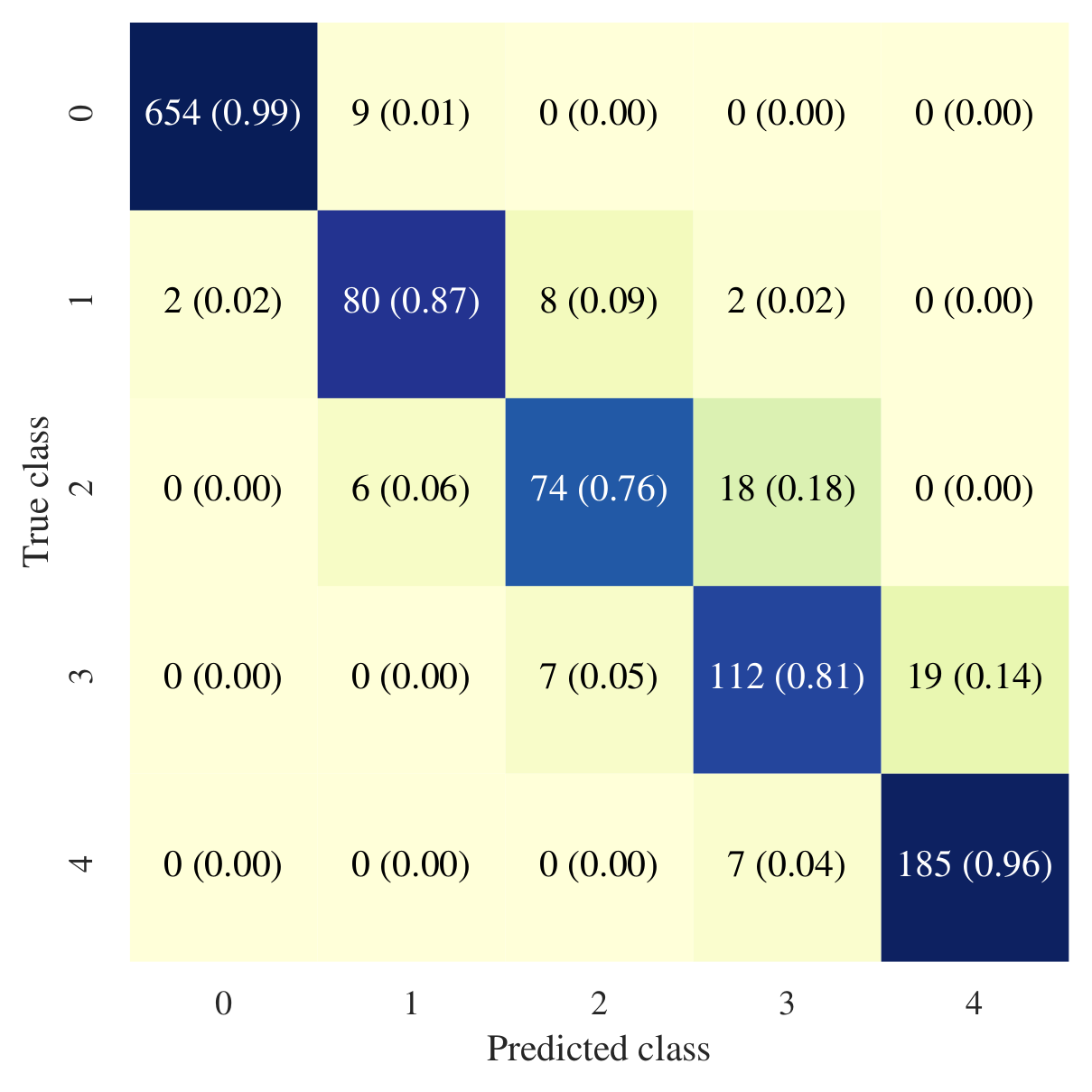}\:
    \centering
    \includegraphics[width=0.4\textwidth]{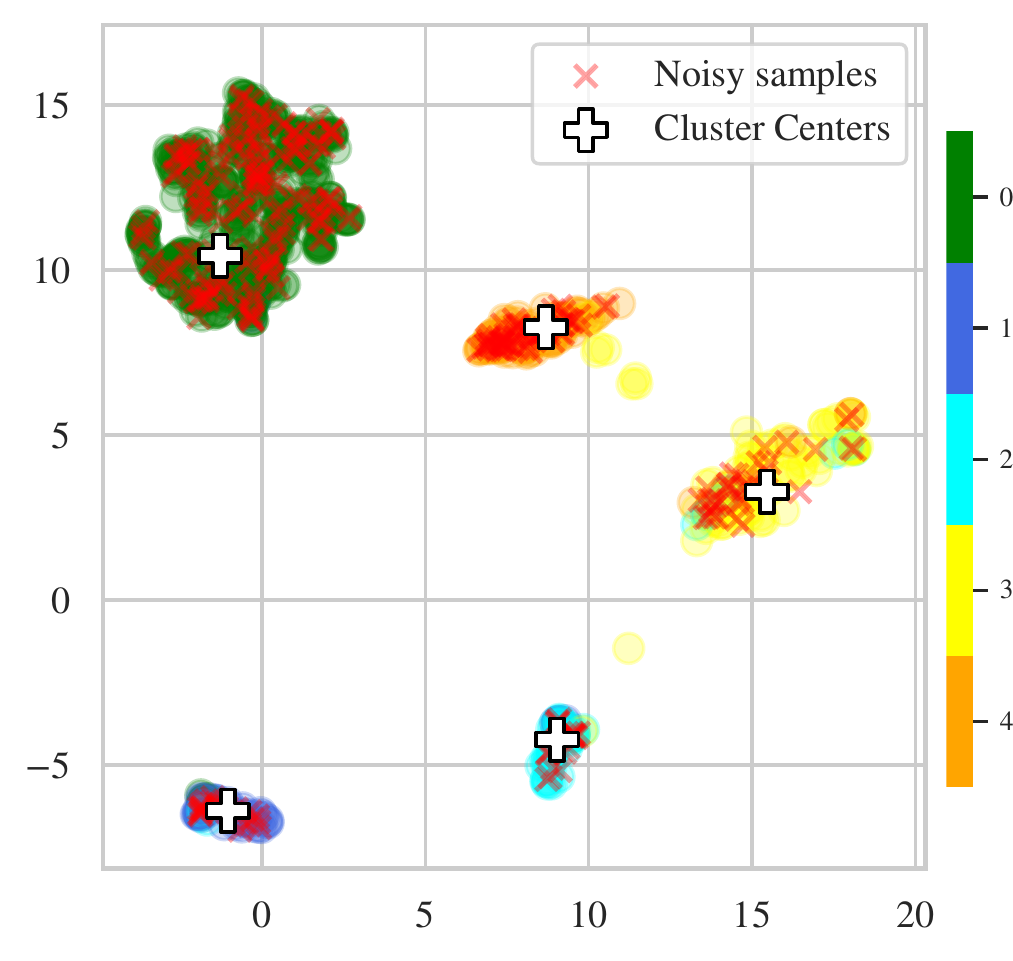}
    \caption{Confusion matrix of the corrected labels (left) and embedding space of the train-set (right) of the CHP power estimation, corrupted with 30\% symmetric noise.}
    \label{fig:cm_umap_symm}
\end{figure}

\begin{figure}[th]
    \centering
    \includegraphics[width=0.4\textwidth]{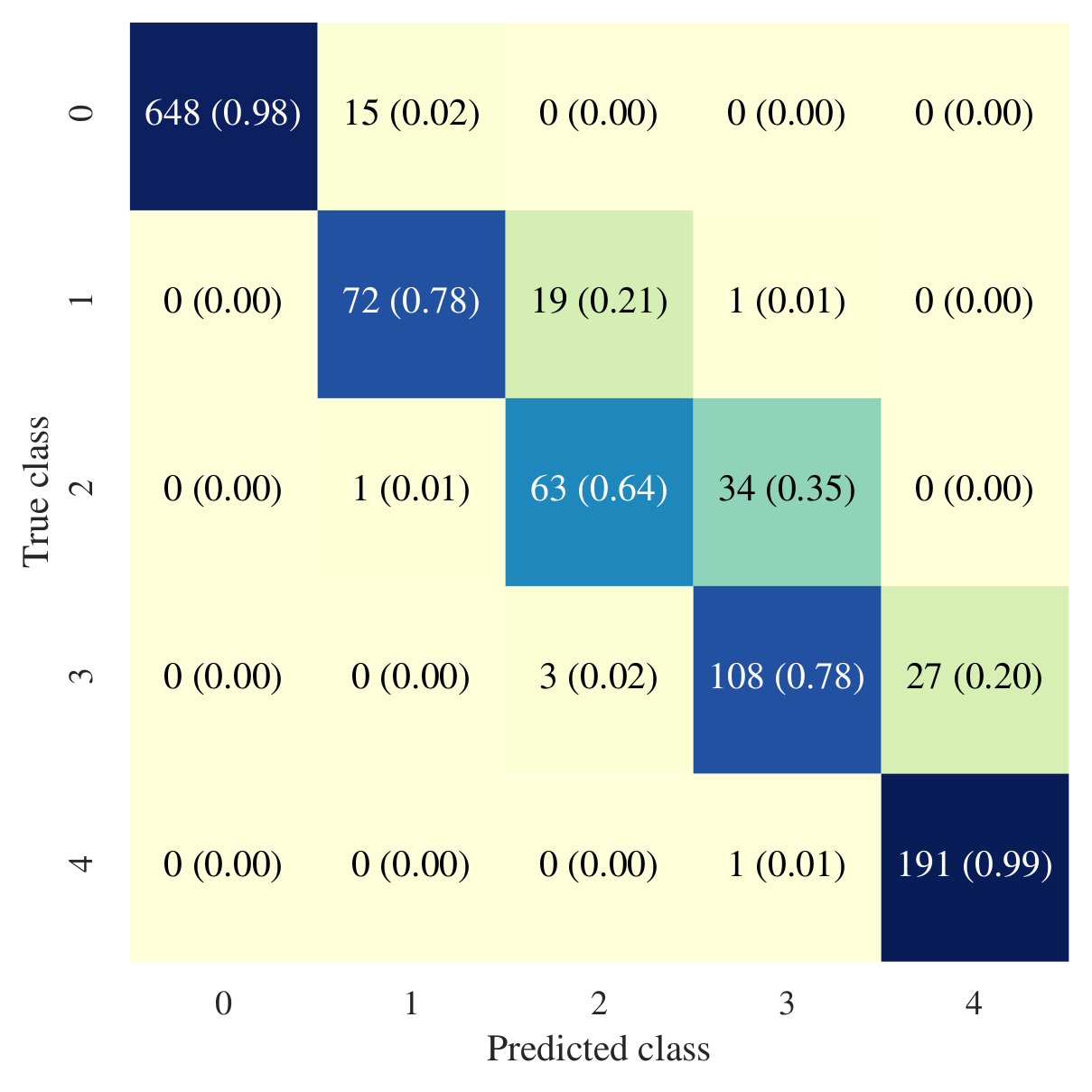}\:
    \centering
    \includegraphics[width=0.4\textwidth]{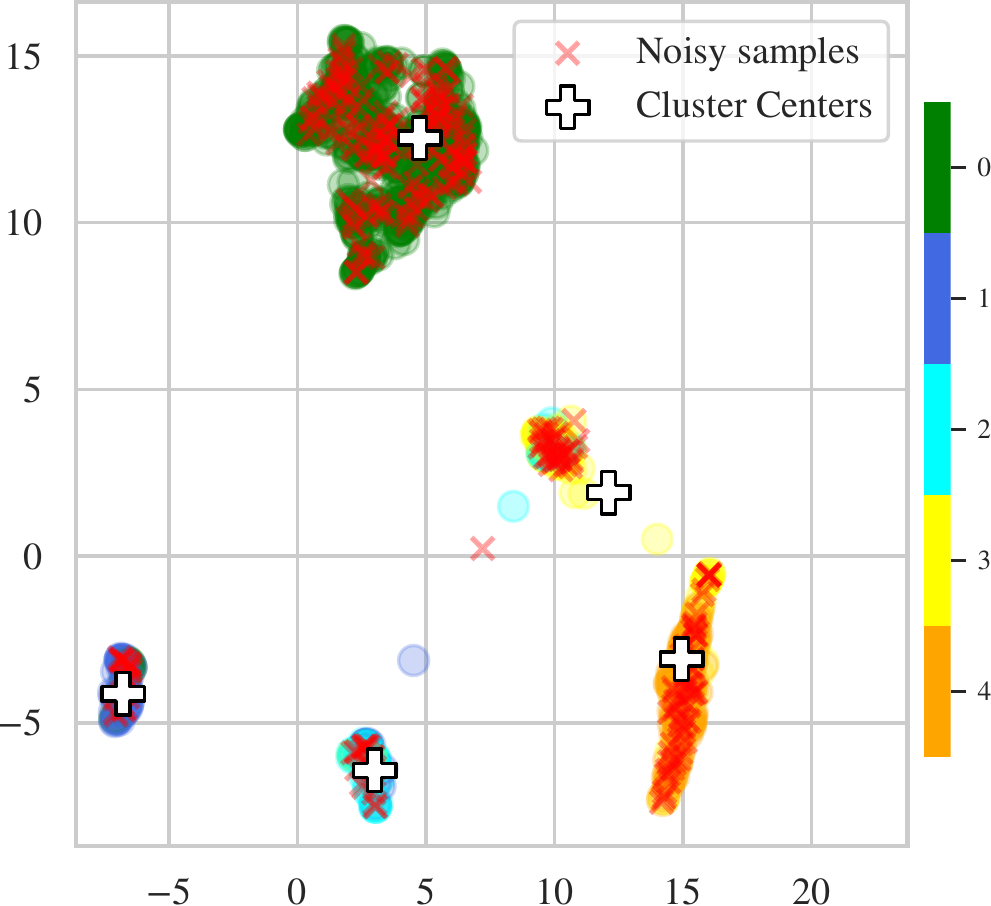}
    \caption{Confusion matrix of the corrected labels (left) and embedding space of the train-set (right) of the CHP power estimation, corrupted with 30\% asymmetric noise.}
    \label{fig:cm_umap_asymm}
\end{figure}

In Fig.\ref{fig:cm_umap_symm} and Fig.~\ref{fig:cm_umap_asymm} we show the confusion matrix of the corrected labels (left) and the embedding space (right) of the \acrshort{method} when training with 30\% of symmetric and asymmetric noise ratio, respectively.
In order to visualize the 32-dimensional embedding space, the UMAP dimensional reduction algorithm \cite{McInnes2018UMAPUM} has been used.
Recall that, for the asymmetric noise type, we circularly shift 30\% of the labels by one i.e $2 \to 3,\; 3 \to 4$. 
This reflects on the errors made on the test set, observed by the confusion matrix in Fig.~\ref{fig:cm_umap_asymm} (left), in particular for the intermediate classes i.e. 1, 2, 3, which are the most challenging for this problem.

\begin{figure}[th]
    \centering
    \includegraphics[width=0.8\textwidth]{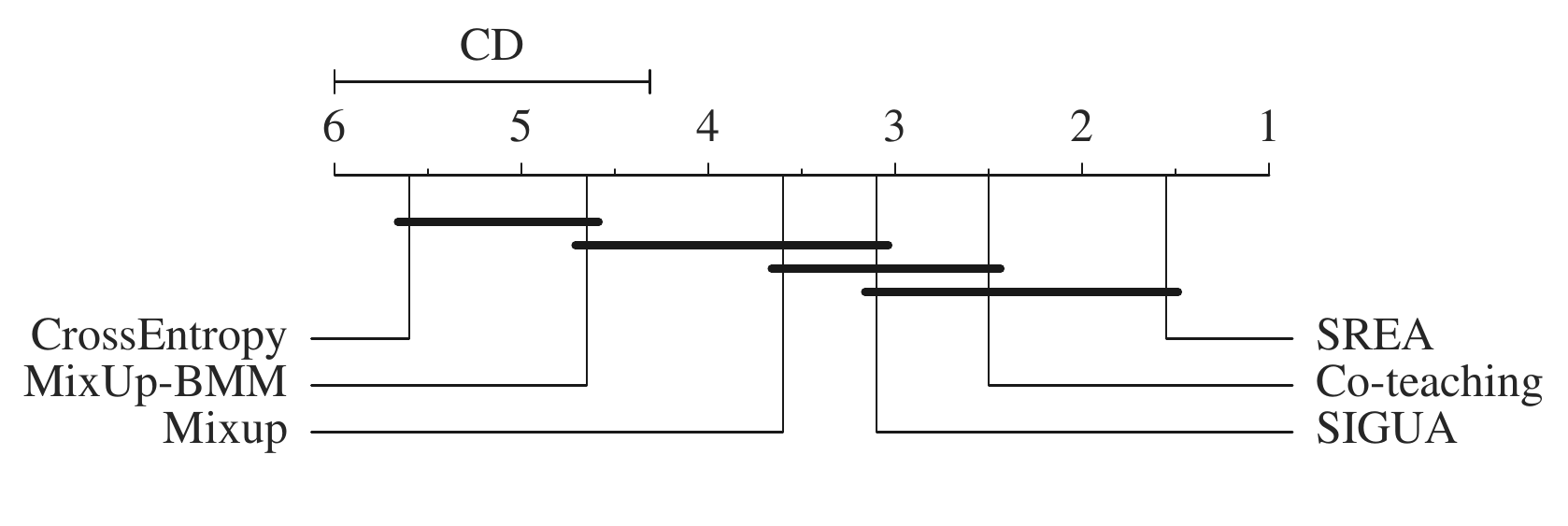}
    \caption{Critical difference diagram showing pairwise statistical difference comparison of \acrshort{method} and SoTA on the CHP dataset corrupted with symmetric label noise.}
    \label{fig:CHP_CD_symm}
\end{figure}

\begin{figure}[th]
    \centering
    \includegraphics[width=0.8\textwidth]{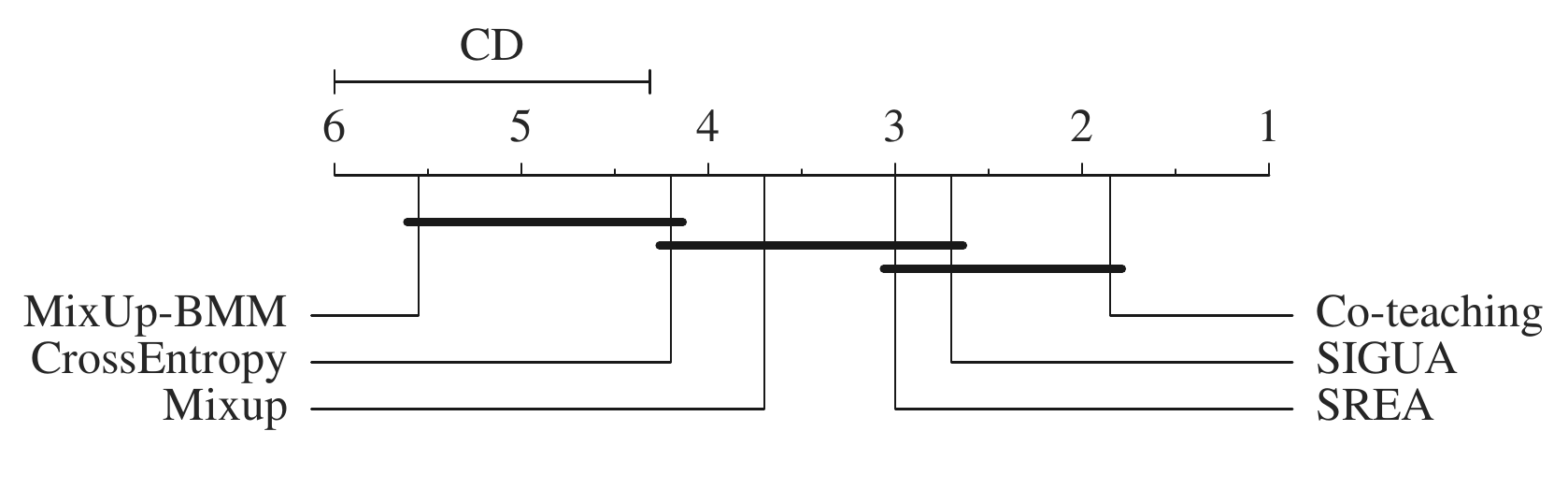}
    \caption{Critical difference diagram showing pairwise statistical difference comparison of \acrshort{method} and SoTA on the CHP dataset corrupted with asymmetric label noise.}
    \label{fig:CHP_CD_asymm}
\end{figure}

\begin{figure}[th]
    \centering
    \includegraphics[width=0.8\textwidth]{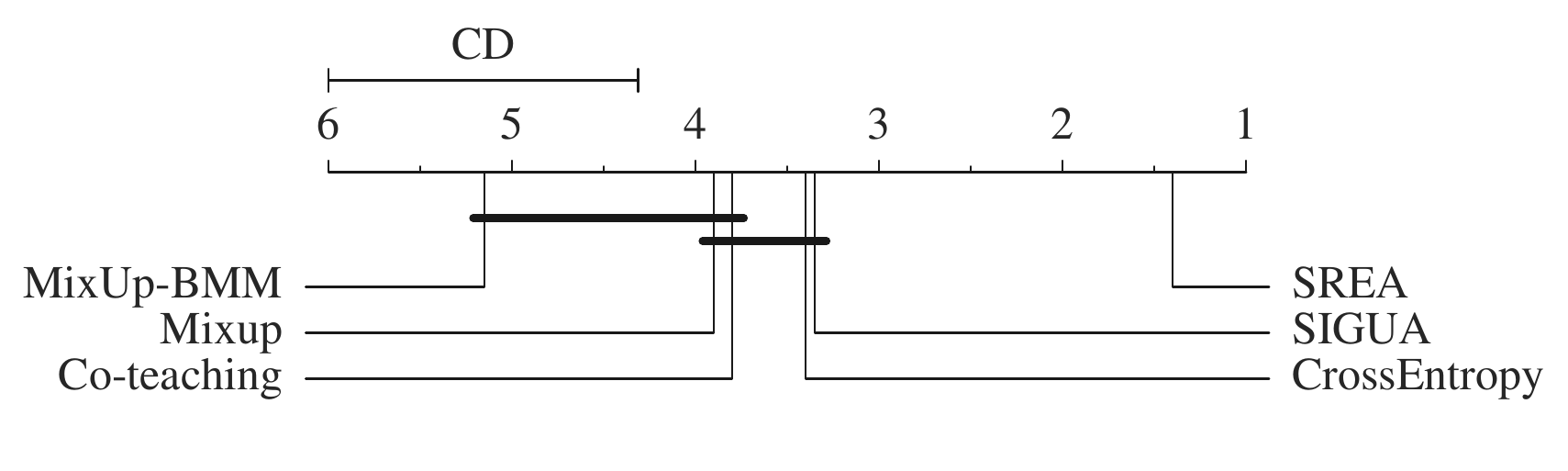}
    \caption{Critical difference diagram showing pairwise statistical difference comparison of \acrshort{method} and SoTA on the CHP dataset corrupted with flip label noise.}
    \label{fig:CHP_CD_flip}
\end{figure}

\begin{figure}[th]
    \centering
    \includegraphics[width=0.8\textwidth]{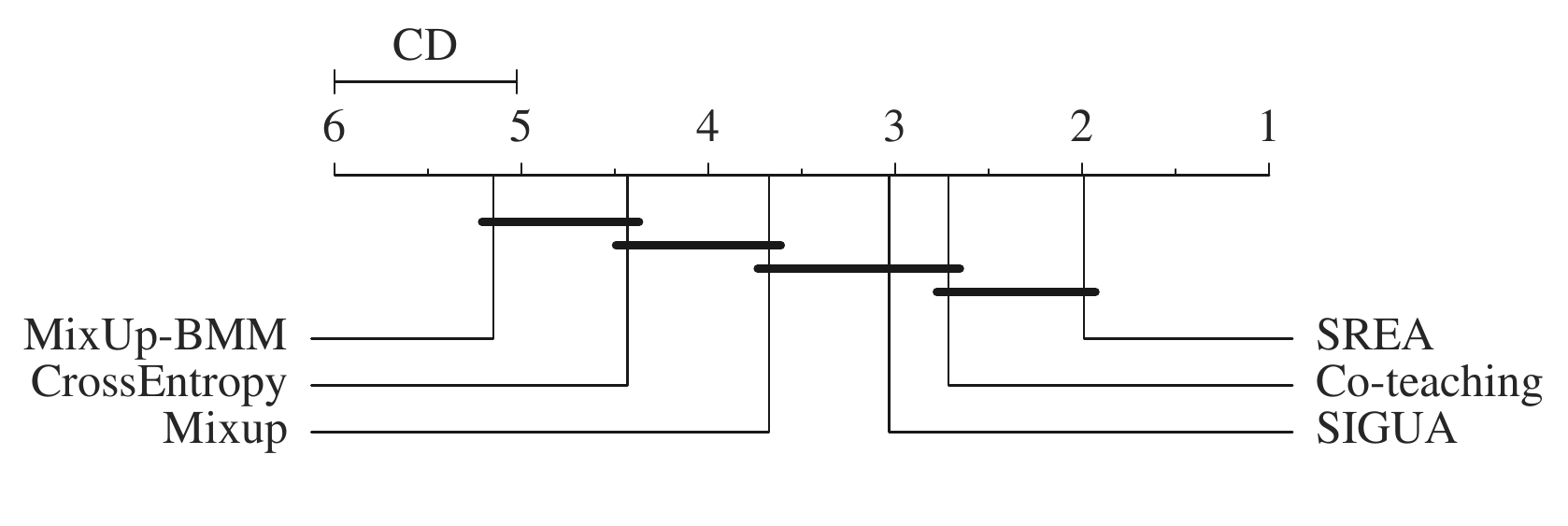}
    \caption{Critical difference diagram showing pairwise statistical difference comparison of \acrshort{method} and SoTA on the CHP dataset corrupted with symmetric, asymmetric and flip label noise.}
    \label{fig:CHP_CD}
\end{figure}

\new{In Fig.\ref{fig:CHP_CD_symm}, \ref{fig:CHP_CD_asymm}, \ref{fig:CHP_CD_flip} is depicted the critical difference diagram 
on the CHP dataset, under presence of symmetric, asymmetric and flip noise respectively.
In Fig.~\ref{fig:CHP_CD} we show the critical difference diagram with all the noise types are combined.}

\begin{figure}[h!]
\begin{subfigure}{\textwidth}
    \centering
    \includegraphics[width=\linewidth]{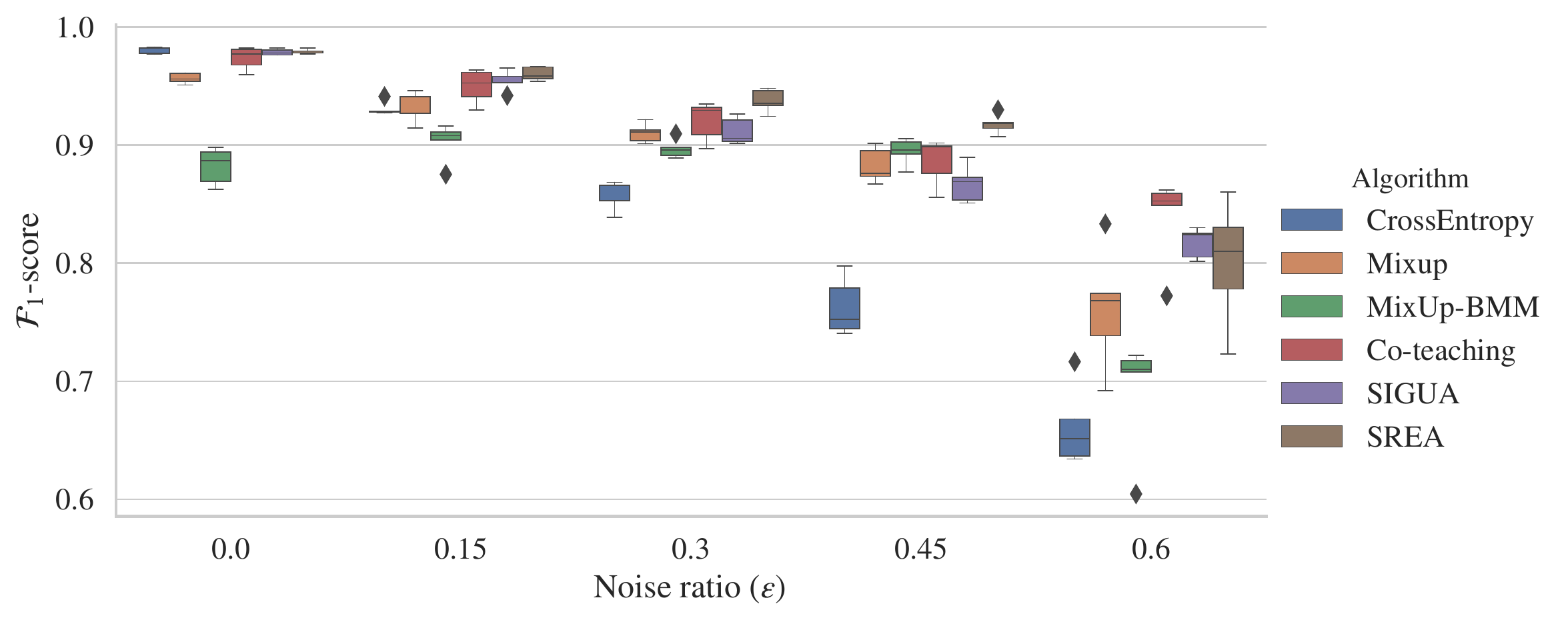}
    \caption{Symmetric noise.}
    \label{fig:CHP_symm}
\end{subfigure}
\begin{subfigure}{\textwidth}
    \centering
    \includegraphics[width=\linewidth]{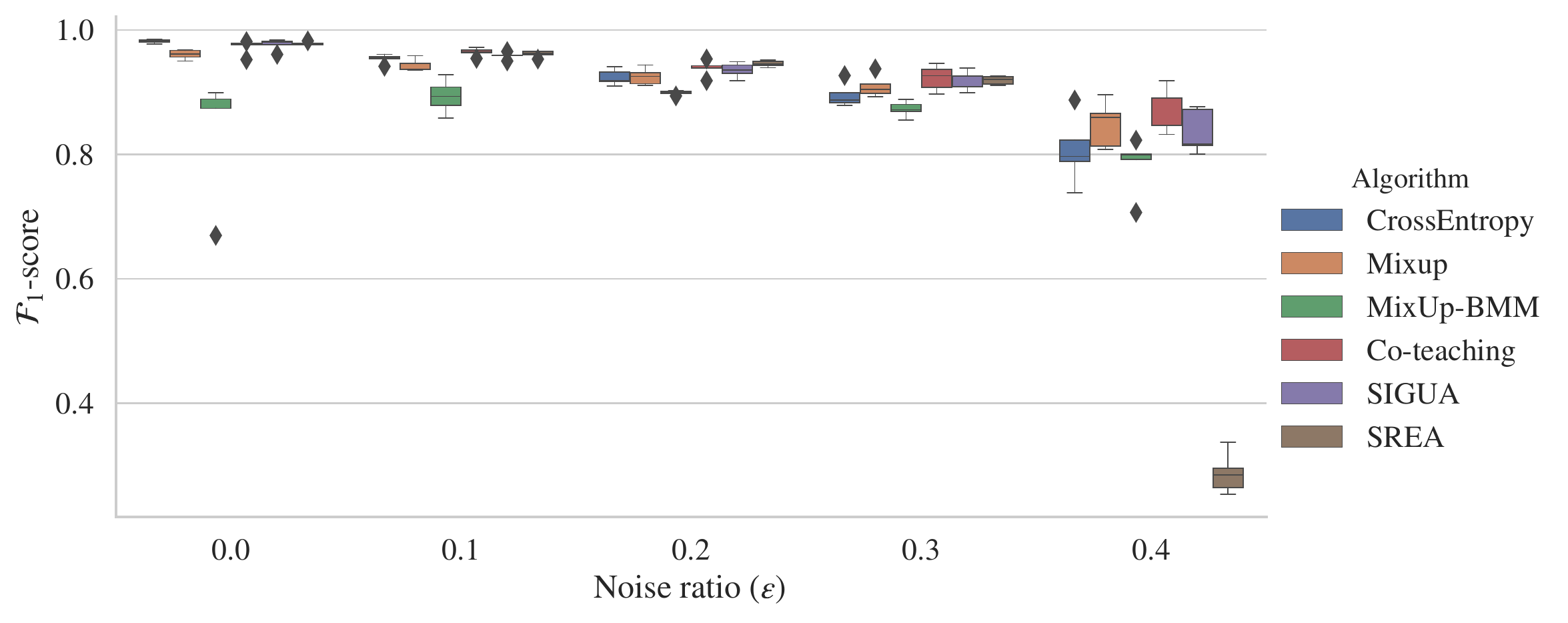}
    \caption{Asymmetric noise.}
    \label{fig:CHP_asymm}
\end{subfigure}
\centering
\begin{subfigure}{\textwidth}
    \centering
    \includegraphics[width=\linewidth]{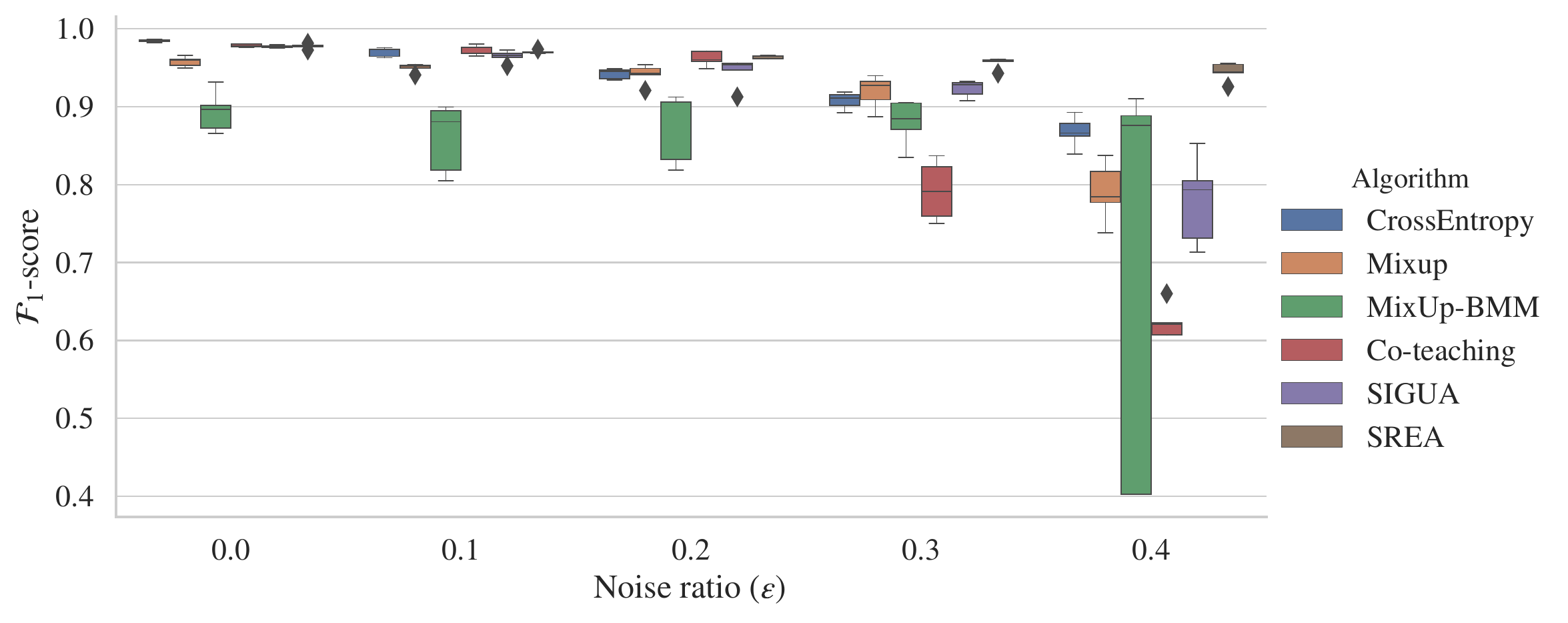}
    \caption{Flip noise.}
    \label{fig:CHP_flip}
\end{subfigure}%
\caption{$\mathcal{F}_1$ test scores of the CHP power estimation when training with different noise type. Each experiment consists of 10 independent runs. }
\label{fig:CHP_results}
\end{figure}

Fig~\ref{fig:CHP_results} depicts the box-plot of the $\mathcal{F}_1$ test score under different types and levels of noise, with the comparison with other SotA algorithm.
It is clearly visible that the proposed \acrshort{method} has comparable (and sometimes better) performance than the other algorithms.

\begin{figure}[H]
    \centering
    \includegraphics[width=\textwidth]{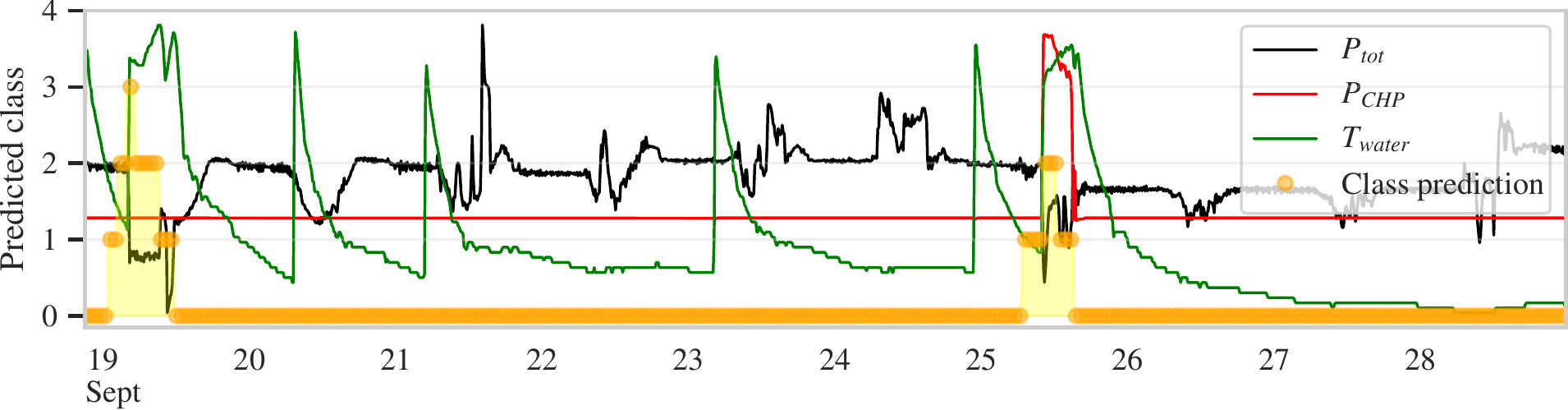}
    \caption{Example of label noise in the CHP dataset corrected with \acrshort{method}.}
    \label{fig:CHP_correct}
\end{figure}

In Fig.~\ref{fig:CHP_correct} we show the results of the \acrshort{method} on the data with the real sensor failure.
We highlight that the method was able to correctly re-label the period of the $P_{CHP}$ sensor failure (night of 19$^{th}$ Sept.) and also detect the other segments where the CHP was active (night of 25$^{th}$ Sept.).
The offset between the detection and the real activation of the machine is due to the sliding windowing operation in the creation of the data.

\clearpage
\section{Ablation studies}

\comment{
\subsection{Input output window size.
}
\andrea{Input - output window size.}
\andrea{I think it is better to remove this. }
\begin{table*}[tbh]
    \centering
    \scriptsize
    \setlength{\tabcolsep}{3pt}
    \begin{tabular*}{\textwidth}{l @{\extracolsep{\fill}} l  c c c c | c c}
    \toprule
    \multirow{2}{*}{$\Delta t_{in}$} & \multirow{2}{*}{$\Delta t_{out}$} & \multicolumn{4}{c |}{\textbf{Symmetric}} & \multicolumn{2}{c}{\textbf{Asymmetric}} \\
    \cmidrule(lr){3-8}
    & & 0 & 15 & 30 & 45 & 20 & 30\\
    \midrule
    60 & 60 & 0.905\rpm0.036 & 0.891\rpm0.043 & 0.866\rpm0.038 & 0.833\rpm0.010 & 0.896\rpm0.031 & 0.752\rpm0.269 \\
    360 & 60 & 0.919\rpm0.034 & 0.892\rpm0.032 & 0.875\rpm0.028 & 0.877\rpm0.034 & 0.864\rpm0.038 & 0.737\rpm0.295\\
    \underline{360} & \underline{360} & \underline{0.979\rpm0.002} & \underline{0.960\rpm0.006} & \underline{0.938\rpm0.010} & \underline{0.918\rpm0.008} & \underline{0.946\rpm0.005} & \underline{0.919\rpm0.007}\\
    \bottomrule
    \end{tabular*}
    \caption{Real-world application. Comparison at varying of intput and target windows. Mean values and standard deviation over 10 runs. Values of $\Delta t_{IN}$ and $\Delta t_{OUT}$ in minutes. \andrea{cause space issues, I have just reported values for low-noise settings. Thats the same reason why I used underline instead of bold to highlight the best values. Is this okay? Should note others? }}
    \label{tab:CHP_window}
\end{table*}
}

\subsection{Input signals.}

In Table~\ref{tab:CHP_exog} we report the ablation studies on the input variables for the estimation of the CHP power level with the algorithm \acrshort{method}, with both symmetric and asymmetric noise.

\begin{table*}[htb]
    \centering
    \scriptsize
    \setlength{\tabcolsep}{2pt}
    \caption{Ablation studies on the input variables for the estimation of CHP power level with \acrshort{method}. The best results per noise type and ratio are underlined.}
    \label{tab:CHP_exog}
    \begin{tabular*}{\textwidth}{l l l @{\extracolsep{\fill}}  c | c c c | c c}
    \toprule
    \multirow{2}{*}{$\bm{P_{tot}}$} & \multirow{2}{*}{$\bm{T_{amb}}$} & \multirow{2}{*}{$\bm{T_{wtr}}$} &\textbf{Noise} & \multicolumn{3}{c|}{\textbf{\textbf{Symmetric}}} & \multicolumn{2}{c}{\textbf{Asymmetric}} \\
    \cmidrule(lr){4-9}
    & & & 0 & 15 & 30 & 45 & 20 & 30\\
    \midrule
    \xmark & \cmark & \xmark & 0.625\rpm0.043 & 0.524\rpm0.036 & 0.530\rpm0.017 & 0.521\rpm0.053 & 0.491\rpm0.041 & 0.531\rpm0.021 \\
    \xmark & \xmark & \cmark & 0.944\rpm0.003 & 0.903\rpm0.006 & 0.853\rpm0.025 & 0.799\rpm0.028 & 0.886\rpm0.004 & 0.822\rpm0.028 \\
    \xmark & \cmark & \cmark & 0.962\rpm0.003 & 0.935\rpm0.008 & 0.901\rpm0.021 & 0.863\rpm0.014 & 0.915\rpm0.015 & 0.869\rpm0.009\\
    \cmark & \xmark & \xmark & 0.938\rpm0.004 & 0.887\rpm0.004 & 0.845\rpm0.010 & 0.793\rpm0.012 & 0.868\rpm0.010 & 0.832\rpm0.011 \\
    \cmark & \cmark & \xmark & 0.955\rpm0.007 & 0.919\rpm0.009 & 0.889\rpm0.005 & 0.835\rpm0.017 & 0.896\rpm0.009 & 0.871\rpm0.009 \\
    \cmark & \xmark & \cmark & 0.974\rpm0.003 & 0.947\rpm0.004 & 0.928\rpm0.007 & 0.885\rpm0.015 & 0.934\rpm0.009 & 0.910\rpm0.009 \\
    \cmark & \cmark & \cmark & \underline{0.978\rpm0.001} & \underline{0.963\rpm0.003} & \underline{0.937\rpm0.007} & \underline{0.914\rpm0.004} & \underline{0.941\rpm0.007} & \underline{0.921\rpm0.010}\\
    \bottomrule
    \end{tabular*}
\end{table*}

\subsection{Hyper-parameter sensitivity}

In this section, we show the $\mathcal{F}_1$ test scores on the complete grid search of the \acrshort{method} hyper-parameters analyzed: $\lambda_{init} \in \{0, 10, 20, 40\}$, $\Delta_{start} \in \{0, 5, 10, 15, 25\}$ and $\Delta_{end} \in \{ 10, 20, 30 \}$.

In Fig.~\ref{fig:CHP_symm} with symmetric noise, in Fig.~\ref{fig:CHP_asymm} with asymmetric noise and in Fig.~\ref{fig:CHP_flip} with flip noise.
In those figures, each row of plots correspond to a different noise ratio, eacg column to a different value for $\Delta_{end}$.
For every subplot, in the x-axis are listed the values for $\lambda_{init}$.

\begin{figure}[th!]
    \centering
    \includegraphics[width=\textwidth]{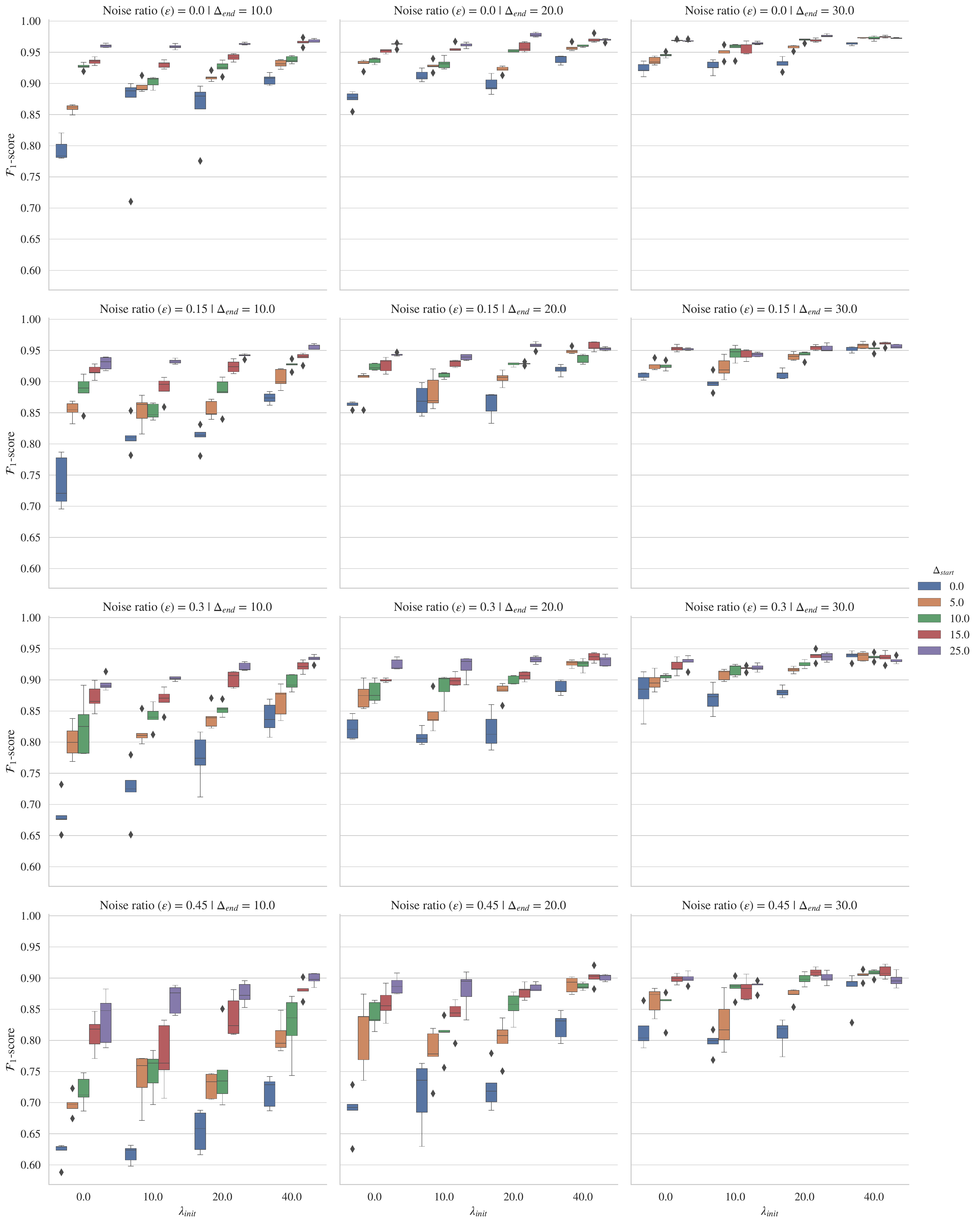}
    \caption{\acrshort{method} hyper-parameter ablation studies on the CHP dataset when training with symmetric noise.}
    \label{fig:CHP_symm}
\end{figure}

\begin{figure}[th!]
    \centering
    \includegraphics[width=\textwidth]{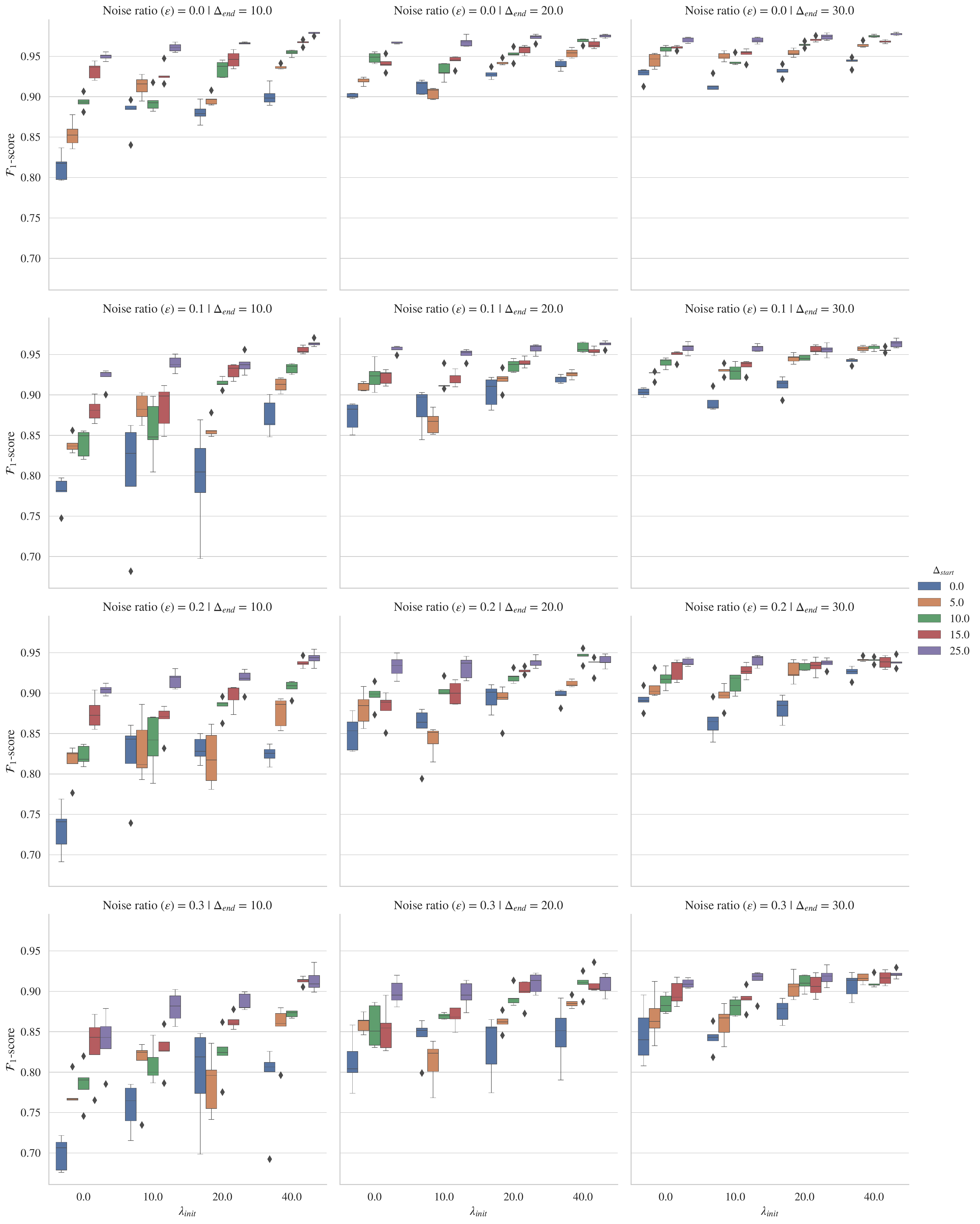}
    \caption{\acrshort{method} hyper-parameter ablation studies on the CHP dataset when training with asymmetric noise.}
    \label{fig:CHP_asymm}
\end{figure}

\begin{figure}[th!]
    \centering
    \includegraphics[width=\textwidth]{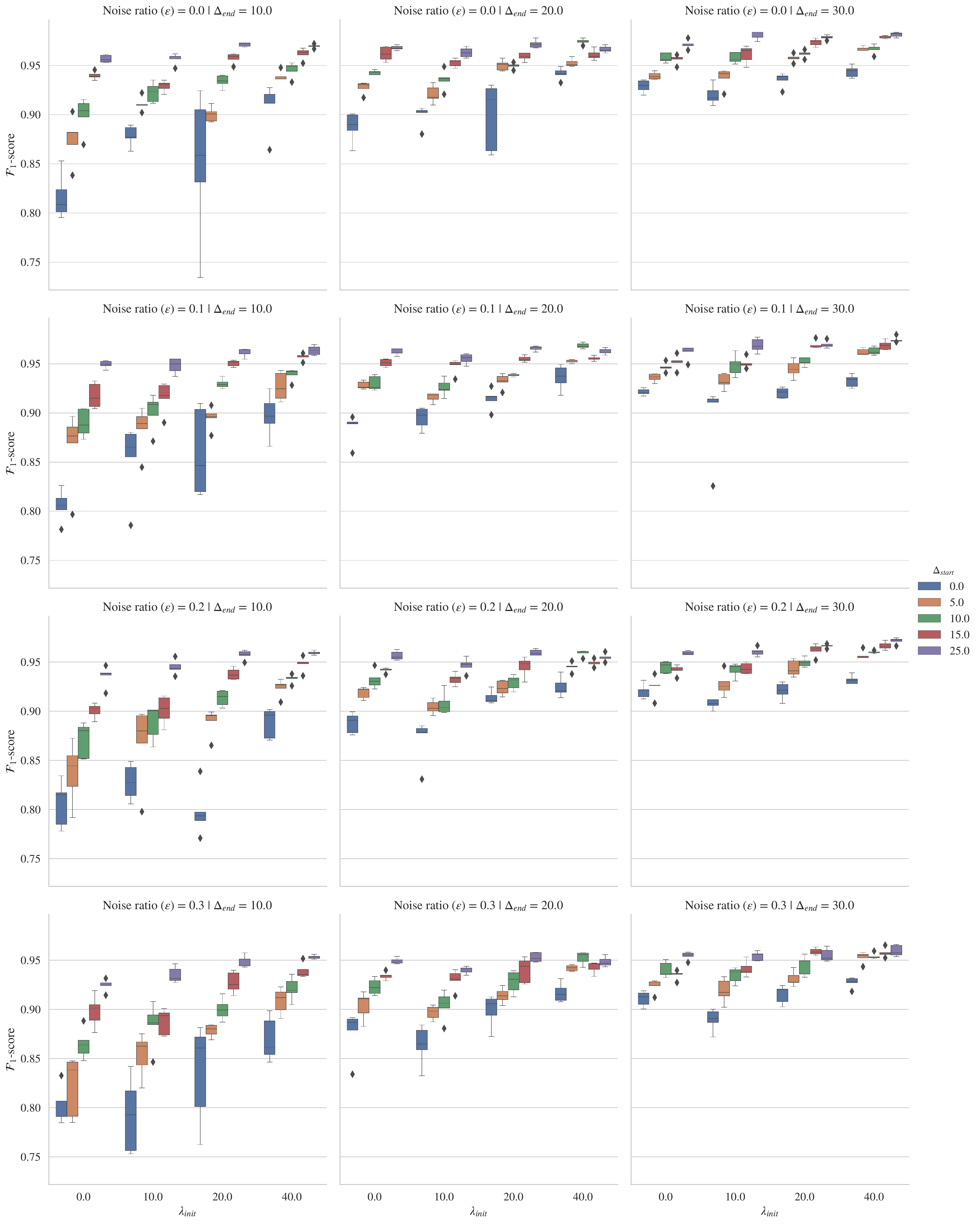}
    \caption{\acrshort{method} hyper-parameter ablation studies on the CHP dataset when training with flip noise.}
    \label{fig:CHP_flip}
\end{figure}

\clearpage
\bibliographystyle{splncs04}
\bibliography{bibliosupplementary.bib}